\pdfoutput=1
\documentclass[sigconf]{acmart}
\AtBeginDocument{%
  \providecommand\BibTeX{{%
    \normalfont B\kern-0.5em{\scshape i\kern-0.25em b}\kern-0.8em\TeX}}}

\setcopyright{none}
\copyrightyear{}
\acmYear{}
\acmDOI{}

\acmConference{Emails: \{penghaozhou,chongzhou,popeyepeng,jeffdu,winfredsun,scorpioguo,garyhuang\}
@tencent.com}
\acmBooktitle{}
\acmPrice{}
\acmISBN{}

\acmSubmissionID{1285}

\usepackage[figure]{algorithm2e}
\usepackage{tcolorbox}

\usepackage{tikz}
\usetikzlibrary{positioning, shapes, arrows}

\usepackage{definitions}

\begin{document}

\title{\nmsname{}: Improving Pedestrian Detection by Nearby Objects Hallucination}


\author{Penghao Zhou \quad Chong Zhou \quad Pai Peng \quad Junlong Du}
\author{Xing Sun \quad Xiaowei Guo \quad Feiyue Huang}
\author{Tencent Youtu Lab}

\renewcommand{\shortauthors}{Zhou, et al.}


\begin{abstract}
    Greedy-NMS inherently raises a dilemma, where a lower NMS threshold will potentially lead to a lower recall rate and a higher threshold introduces more false positives. This problem is more severe in pedestrian detection because the instance density varies more intensively. However, previous works on NMS don't consider or vaguely consider the factor of the existent of nearby pedestrians. Thus, we propose \heatmapname{} (\heatmapnameshort{}), which pinpoints the objects nearby each proposal with a Gaussian distribution, together with \nmsname{}, which dynamically eases the suppression for the space that might contain other objects with a high likelihood. Compared to Greedy-NMS, our method, as the state-of-the-art, improves by $3.9\%$ AP, $5.1\%$ Recall, and $0.8\%$ MR\textsuperscript{-2} on CrowdHuman to $89.0\%$ AP and $92.9\%$ Recall, and $43.9\%$ MR\textsuperscript{-2} respectively.
    
\end{abstract}




\maketitle
\section{Introduction}


Non-maximum Suppression (NMS) is widely used in proposal-based object detectors~\cite{yolo-v1, yolo-v2, yolo-v3, ssd, dssd, rcnn, fast-rcnn, faster-rcnn, fpn, mask-rcnn, cascade-rcnn, rfcn, deformable, focal-loss}, as the post processing step to eliminate the redundant detections. Ideally, the proposal with the maximum score should suppress and only suppress all the other proposals of the same object. However, NMS distinguishes objects solely by a universal Intersection over Union (IoU) threshold. That is, if two proposals have an IoU above the pre-defined threshold, they will be considered as \textit{detecting the same object} and one of them will be eliminated as the duplicate.

\begin{figure}[t]
\begin{center}
\includegraphics[width=1.0\linewidth]{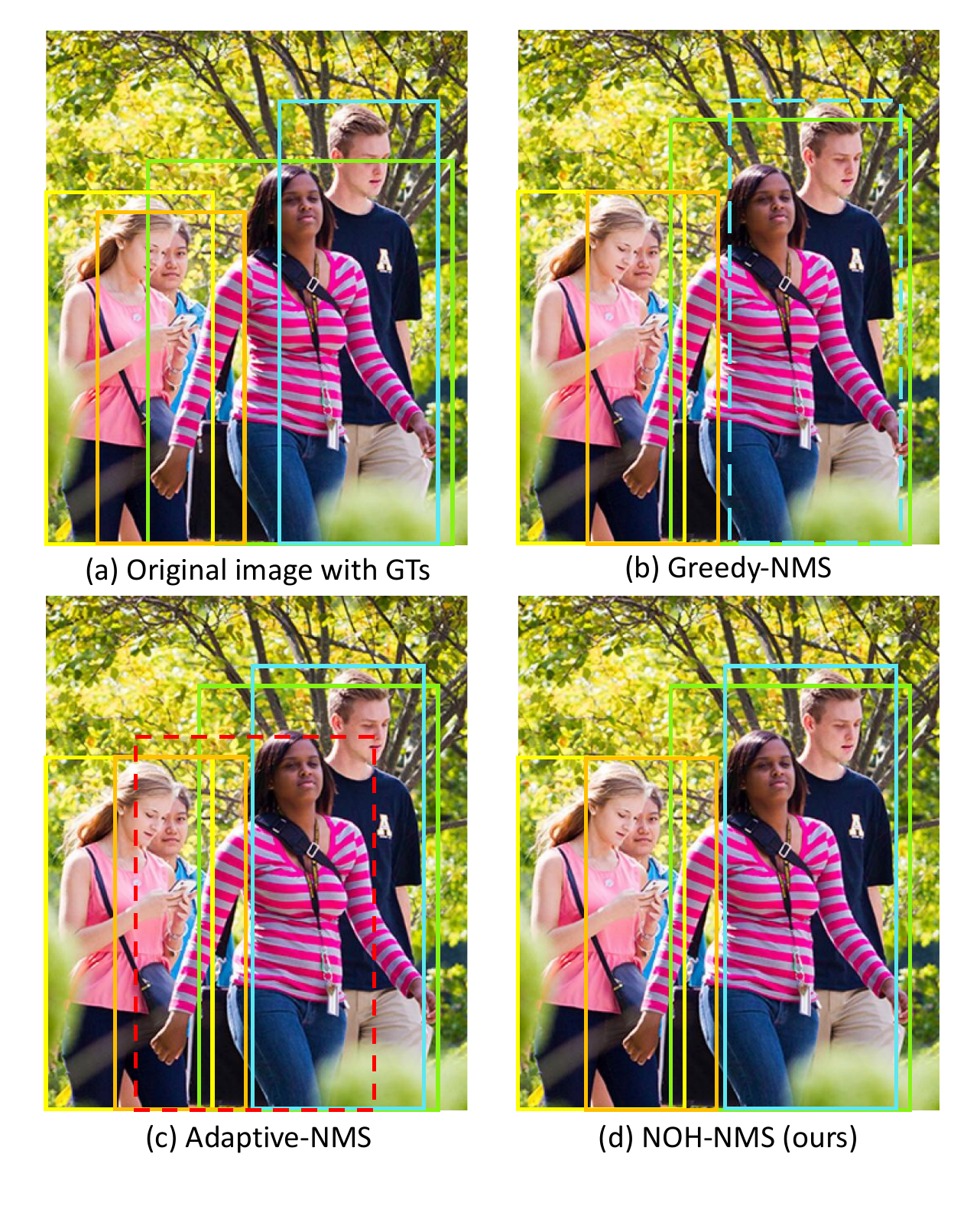}
\end{center}
\vspace{-0.3cm}
\caption{\textbf{Comparison among various NMS methods} {\normalfont The blue dotted box in (b) shows the mistakenly suppressed detection, which is caused by the NMS threshold dilemma in Greedy-NMS. The red dotted box in (c) highlights the false positive introduced by Adaptive-NMS as it is unable to pinpoint the overlapping areas. In order to recall the detection of the boy and suppress the red box, adapting the IoU threshold is not enough since $iou(box_{red}, box_{green}) < iou(box_{blue}, box_{green})$, and \heatmapnameshort{} is designed for filling this gap.}}
\vspace{-0.3cm}
\label{fig:intro}
\end{figure}
This scheme works fine in generic object detection task. However, it raises a dilemma in pedestrian detection task where the object density varies a lot, making it infeasible to find a perfect universal IoU threshold as a higher threshold fits for the regions with higher density and the less crowded regions desire a lower threshold (See Fig.~\ref{fig:intro}).

Previous work tries to address this issue of the rigid NMS threshold. Soft-NMS~\cite{soft-nms} proposes to degrade the score of nearby highly overlapped proposals instead of eliminating them, but just like Greedy-NMS, it still blindly penalizes the highly overlapped boxes. Adaptive-NMS~\cite{adaptive-nms} suggests directly predicting a proper NMS threshold for each proposal. However, even though the proposal can sense the density of the nearby objects, it is not aware of the locations and spread of the crowded regions, which results in a new dilemma, as shown in Fig.~\ref{fig:intro}, where the left to the proposal is not dense at all and the right is rather crowded.

Thus, to tackle this problem, we propose \heatmapname{} (\heatmapnameshort{}) and \nmsname{}. Our key observation is, in a crowded scene, the visual information inside a bounding box of one pedestrian will mostly contain the cues of the locations and sizes of other pedestrians. Therefore, we design \heatmapnameshort{}, which hallucinates the objects nearby each proposal based on the Region-of-Interest (RoI) feature and represents the hallucination with a Gaussian distribution. Furthermore, we propose \nmsname{} to perform a novel NMS strategy leveraging the Gaussian distribution.

The proposed \heatmapnameshort{} and \nmsname{} can be integrated naturally into both one-stage and two-stage object detectors with marginal computation cost and acquire no more extra annotations other than the full-body bounding boxes during training.

To evaluate the effectiveness of our method, we have conducted quantitative and qualitative experiments on CityPersons~\cite{citypersons} and CrowdHuman~\cite{crowdhuman} datasets (see Sec. \ref{sec:experiments}). As a result, we achieve state-of-the-art performance with $89.0\%$ AP, $92.9\%$ Recall, $43.9\%$ \mr{} on CrowdHuman, and $10.8\%$ \mr{} on CityPersons.

Our contributions can be summarized as follow:
\begin{itemize}
  \item We propose \nmsname{}, which is aware of the existence of other nearby objects when performing the suppression, to address the rigid NMS threshold problem in pedestrian detection.
  \item We design \heatmapnameshort{} to pinpoint the objects nearby each proposal with a Gaussian distribution.
  \item Our method achieves state-of-the-art performance on CityPersons and Crowdhuman with negligible overhead.
\end{itemize}

\section{Related Work}
Over the past decade, deep convolutional neural networks (CNNs) have made great strides in image recognition \cite{resnet}. To adapt an image classifier into an object detector, the current common practice, called proposal-based object detector, leverages sliding window to densely predict, for each proposal, a set of category confidence scores and proposal refinement coefficients. These refined proposals are then fed into the NMS algorithm to get rid of the redundant detections. According to different strategies to generate the proposals, proposal-based object detectors can be classified into one-stage, where proposals are pre-defined anchors, and two-stage, where Region Proposal Networks (RPNs) are used for proposal generation. In addition, great progress has been made in multiple scaling~\cite{fpn, panet}, learnable anchors~\cite{guidedanchor, metaanchor}, deformable feature sampling~\cite{deformable, deformablev2}, etc.

Even though state-of-the-art generic object detectors show promising performance on benchmark datasets, such as COCO~\cite{coco} and Pascal VOC~\cite{pascal}, it is non-trivial to adapt them into the pedestrian detection task, because the occlusion is much more severe and frequent in pedestrian detection datasets.

Occlusion can be divided into two categories, namely inter-class occlusion and intra-class occlusion. In intra-class occlusion scenarios, the pedestrian is occluded by other pedestrians. And the inter-class occlusion results in the partially visible feature of pedestrians mixed with the feature of background objects.

To address the problem of inter-class occlusion, some algorithms \cite{bibox,mgan,orcnn} seek to leverage the annotated visible bounding box (VBB). \cite{bibox} introduces a visible part estimation branch and a new training sample selecting strategy assisted by VBB. OR-CNN~\cite{orcnn} exploits the topological structure of the pedestrian with visibility prediction for occluded pedestrian detection. To emphasize on visible pedestrian regions during feature extraction, MGAN~\cite{mgan} proposes an attention module supervised by VBB.

In intra-class occlusion scenarios, the pedestrian is occluded by other pedestrians, which occurs frequently in the crowd scene. The heavily occluded between pedestrians confuses the models as it's hard to distinguish instance boundaries. To alleviate this problem, OR-CNN~\cite{orcnn} designs aggregation loss to enforce generating more compact bounding boxes. In addition, RepLoss~\cite{repulsion-loss} proposes a novel repulsion loss to prevent the proposal from shifting to surrounding objects. 

Though OR-CNN~\cite{orcnn} and RepLoss~\cite{repulsion-loss} successfully ease the localization problem in the crowded scenes, there still exists an even worse issue during the post-processing stage. In the post-processing stage, Non-maximum Suppression (NMS) is wildly used to suppress false positive proposals (i.e., the redundant pedestrian proposals belong to the same identity). However, NMS may also suppress true positive proposals (i.e., the highly overlapped pedestrian proposals belong to different identities). Therefore, a lower threshold leads to a lower Recall while a higher threshold results in lower precision.

To address this dilemma, ~\cite{soft-nms} proposes Soft-NMS to replace the elimination operation with decaying the detection scores according to the IoU. And~\cite{doubleanchor,jointdet} suggest using additional annotated head bounding boxes to solve the problem of NMS in a crowd, as the head parts usually suffer less from occlusion. More recently, Adaptive-NMS~\cite{adaptive-nms} proposes to predict the adaptive IoU threshold in NMS for each proposal. It aims at predicting a higher NMS threshold if the objects gather together and occlude each other, and predicting lower NMS threshold if the objects are sparse. However, even though Adaptive-NMS could predict accurate density for each proposal, a density scalar is not enough to precisely express the spatial locations of the crowded areas. In other words, the proposal is capable of sensing how crowded its surrounding is, but cannot tell if the area to its left is more crowded than the area to its right. As a result, Adaptive-NMS gets stuck into a new dilemma when different spatial locations to one object desire different IoU thresholds, as shown in Fig.\ref{fig:intro}.

We observe this inflexibility in Adaptive-NMS and thus propose \heatmapnameshort{} and \nmsname{} to address this problem. Specifically, for \heatmapnameshort{}{}, we design a mini 2-fc branch to predict, for each proposal, not only a density scalar but also a Gaussian distribution which highlights the surrounding objects. In addition, our \nmsname{} leverages the output from \heatmapnameshort{} as the auxiliary information, together with the normal NMS input (detection boxes with class confidence), to perform a nearby-objects-aware NMS.

\section{Our Method}
In this section, we first briefly recap the previous NMS algorithms (Sec.~\ref{sec:background}). Then we propose our \nmsname{} which integrates the nearby-objects distribution into the NMS pipeline (Sec.~\ref{sec:nms}). In addition, we illustrate how our \heatmapnameshort{} module learns to predict the nearby-objects distribution just from the box-level supervision (Sec.~\ref{sec:heatmap}). Finally, we compare our method with the state-of-the-art NMS counterparts in with visualization (Sec.~\ref{sec:compare}).

\begin{algorithm}
    \SetKwInOut{Input}{Input}
    \SetAlgoLined
    \Input{
        $
            \mathcal{B} = {b_1,\dots,b_N}, 
            \mathcal{S} = {s_1,\dots,s_N},
        $
        \newline
        $
            \mathcal{D} = {d_1,\dots,d_N},
            \mathcal{P} = {p_1,\dots,p_N},
            N_t
        $
        \newline
        $\mathcal{B}$ is the list of initial detection boxes \newline
        $\mathcal{S}$ contains corresponding detection scores
        \newline
        \textcolor{mygreen}{
            $\mathcal{D}$ contains corresponding detection densities
            \newline
            $\mathcal{P}$ contains the parameters of nearby-objects distribution of corresponding detection
        }
        \newline
        $N_t$ is the NMS threshold
        }
    \Begin{
        $\mathcal{F} \leftarrow \{\}$ \\
        \While{$\mathcal{B} \ne empty$}{
            $m \leftarrow \text{argmax } \mathcal{S}$\\
            $\mathcal{M} \leftarrow b_m$\\
            $\mathcal{F} \leftarrow \mathcal{F} \bigcup \mathcal{M};
            \mathcal{B} \leftarrow \mathcal{B} - \mathcal{M}$\\
            \For{$b_i \in \mathcal{B}$}{
                \If{iou$(\mathcal{M}, b_i) \geq N_t$}{
                    \begin{tcolorbox}[
                        standard jigsaw,
                        opacityback=0,
                        colframe=myred,
                        text width=150pt,
                        boxsep=3pt,
                        left=0pt,right=0pt,top=0pt,bottom=0pt,
                    ]
                    \textcolor{myred}{
                    $
                        \mathcal{B} \leftarrow \mathcal{B} - b_i;
                        \mathcal{S} \leftarrow \mathcal{S} - s_i;
                    $}
                    \begin{flushright}
                        Greedy-NMS
                    \end{flushright}
                    \end{tcolorbox}
                    
                    \begin{tcolorbox}[
                        standard jigsaw,
                        opacityback=0,
                        colframe=mygreen,
                        text width=150pt,
                        boxsep=3pt,
                        left=0pt,right=0pt,top=0pt,bottom=0pt,
                    ]
                    \textcolor{mygreen}{
                    $
                        s_i \leftarrow s_i \cdot f(\m, b_i, d_\m, p_\m);
                    $}
                    \begin{flushright}
                        \nmsname{}
                    \end{flushright}
                    \end{tcolorbox}
                }
            }
        }
        \Return{$\mathcal{F}, \mathcal{S}$}
    }
    \caption{\textbf{Algorithm pseudo code} {\normalfont \nmsname{} replaces the pruning step (highlighted in red) in Greedy-NMS with a nearby-objects-aware re-scoring function (marked with green).}}
    \label{fig:algorithm}
\end{algorithm}

\subsection{Background} 
\label{sec:background}
A proposal-based object detection framework consists of the following five stages: (1) extracting full-image-level feature; (2) generating bounding box proposals; (3) extracting proposal-region-level feature; (4) performing classification and box regression for each proposal; (5) removing redundant detections. In this pipeline, the proposals are usually densely arranged and there is no punishment if two or more detections are detecting the same object. Thus, prior to stage 5, it is rather common that one object area is occupied with multiple detections whereas only one of them counts towards true positive, and the rest are considered as false positive. 

To avoid the aforementioned problem, Greedy-NMS selects the detection with the maximum score $\m$ and eliminates its surrounding inferior detections whose IoU with $\m$ is above certain threshold $N_t$, and then repeats this pruning process with the next best detection, as shown in Fig.~\ref{fig:algorithm}. The pruning step, as the core of the NMS algorithm, can be formulated into a re-scoring function as follow:
\begin{equation}
    s_i=\left\{
        \begin{array}{ll}
            s_i,
            &\text{iou}(\m, b_i) < N_t 
            \\
            0,
            &\text{iou}(\m, b_i) \geq N_t
        \end{array}
        ,
        \right.
    \label{eq:greedy-nms}
\end{equation}

where $s_i$ and $b_i$ denote the confidence score and bounding box coefficients of the inferior detections. $b_i$ will be either left unmodified or completely removed depending solely on its IoU with $\mathcal{M}$. This introduces two problems. (1) The consequence is too extreme and IoU, as the only metric, is not robust enough, which makes the performance very sensitive to the choice of the NMS threshold. E.g., when $N_t$ is set to $0.5$, detection $b_i$ will be eliminated if iou$(\m, b_i)$ equals to $0.51$, however, with a slight perturbation, iou$(\m, b_i)$ could become $0.49$, which makes $b_i$ survive. (2) There is no such NMS threshold that makes everyone happy. E.g., an image occupied with $100$ objects might desire $0.3$ as the threshold, while it is not suitable for the image with a single object.

In response to the first problem, Soft-NMS softens the consequence by gradually decaying the score of the overlapped detections instead of eliminating them. Below shows its re-scoring function:
\begin{equation}
    s_i=\left\{
        \begin{array}{ll}
            s_i,
            &\text{iou}(\m, b_i) < N_t 
            \\
            s_i \cdot f(\m, b_i),
            &\text{iou}(\m, b_i) \geq N_t
        \end{array}
        ,
        \right.
    \label{eq:soft-nms}
\end{equation}

where decaying function $f$ is chosen to be:
\begin{equation}
    f(\m, b_i) = 
    1 - \text{iou}(\m, b_i)
    \;\; \textbf{or} \;\;
    \text{exp}(-\text{iou}(\m, b_i)^2 / \sigma)
    \label{eq:soft-nms-f}
\end{equation}

For the second problem, Adaptive-NMS customizes an NMS IoU threshold for each proposal and follows the design of Greedy-NMS except now the IoU threshold $N_{\m}$ varies with the current best detection $\m$. Their strategy can be formulated as:
\begin{gather}
    N_\m := \text{max}(N_t, d_\m),\\
    s_i=\left\{
        \begin{array}{ll}
            s_i,
            &\text{iou}(\m, b_i) < N_\m
            \\
            0,
            &\text{iou}(\m, b_i) \geq N_\m
        \end{array}
        ,
        \right.
    \label{eq:adaptive-nms-old}
\end{gather}

where $d_\m$ is the density prediction of proposal $\m$.

As we carefully re-visit Adaptive-NMS, we find that due to the \textit{maximum} function, Adaptive-NMS can be re-written into a super case of Soft-NMS:
\begin{equation}
    s_i=\left\{
        \begin{array}{ll}
            s_i,
            &\text{iou}(\m, b_i) < N_t 
            \\
            s_i \cdot f(\m, b_i, d_\m),
            &\text{iou}(\m, b_i) \geq N_t
        \end{array}
        ,
        \right.
    \label{eq:adaptive-nms}
\end{equation}
where
\begin{equation}
    f(\m, b_i, d_\m)=\left\{
        \begin{array}{ll}
            1,
            &\text{iou}(\m, b_i) < d_\m
            \\
            0,
            &\text{iou}(\m, b_i) \geq d_\m
        \end{array}
        \right.
    \label{eq:adaptive-nms-f}
\end{equation}

As shown in Eq.~\ref{eq:soft-nms} and Eq.~\ref{eq:adaptive-nms}, compared to Greedy-NMS, Soft-NMS adds the location of $b_i$ into consideration when suppressing $b_i$ and Adaptive-NMS further considers the density of $\m$. However, both of them cannot accurately distinguish whether $b_i$ is detecting a nearby object or $b_i$ is a false positive. Although equipped with density prediction, Adaptive-NMS still cannot tell where the objects around $\m$ are, let alone Soft-NMS.

\subsection{\nmsname{}}
\label{sec:nms}
The key idea of our \nmsname{} is to introduce the nearby-objects distribution $P_{p_\m}$ into the NMS pipeline, where $p_\m$ denotes its parameters. The nearby-objects distribution could be obtained by any probability distribution functions (PDFs), and we will cover our choice of generating $P_{p_\m}$ in Sec.~\ref{sec:heatmap}. Note that, in the pedestrian detection task, the only object category we care is human, therefore the \textit{nearby objects} refer to \textit{nearby pedestrians} mostly in this paper. However, our method can also be used in other tasks where the \textit{nearby objects} won't be limited to humans only.

\begin{figure*}[t]
\begin{center}
\includegraphics[width=1.0\linewidth]{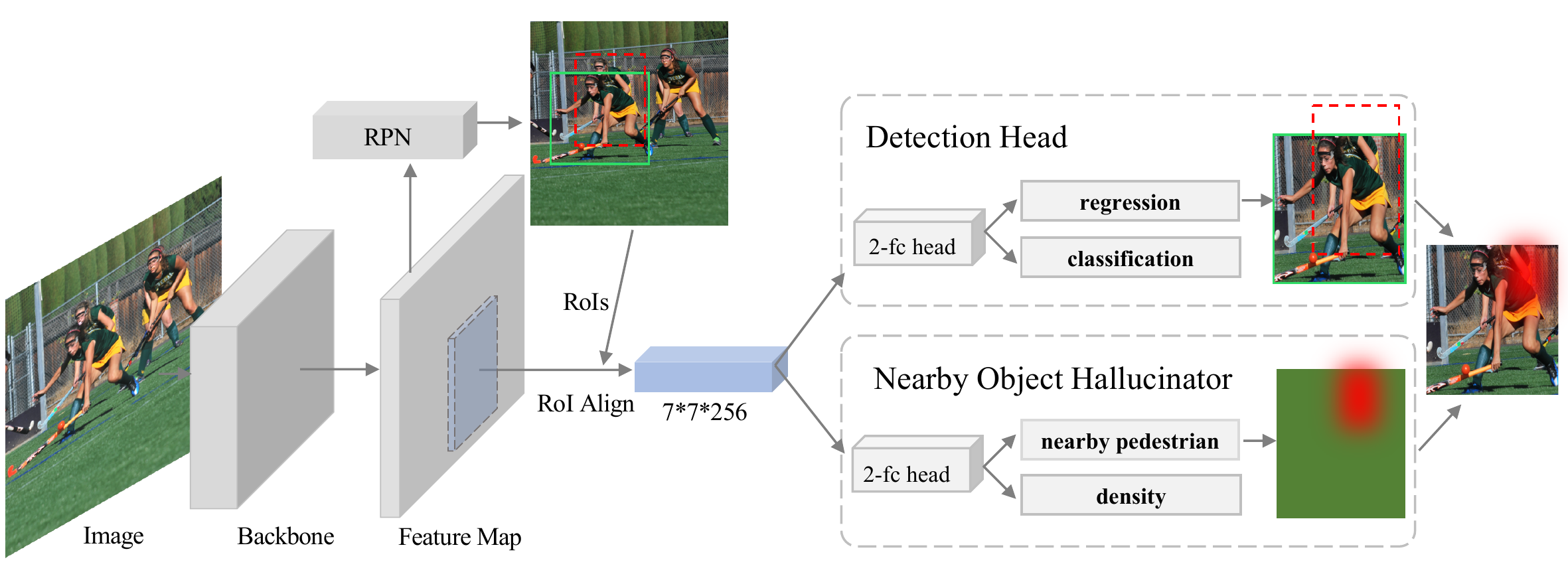}
\end{center}
\caption{\textbf{Architecture} {\normalfont The illustration of integrating \heatmapname{} (\heatmapnameshort{}) into the two-stage object detector, such as Faster-RCNN~\cite{faster-rcnn}. Note that our \heatmapnameshort{} can fit in single-stage object detectors as well by placing the \heatmapnameshort{} branch in parallel with the detection head. In this example, the lady at the front left is highly overlapped with the lady behind her, and our NOH pinpoints the location and shape of the lady behind so that the detection of her won't be mistakenly suppressed whereas other false positives will be eliminated.}}
\label{fig:architecture}
\end{figure*}

\nmsname{} consists of two components, namely overlap detector and NOH-Suppressor. 

\textbf{Overlap Detector} Since our assumption is that the bounding box area of one pedestrian will mostly contain the cues of other pedestrians, we need to first rule out the cases where the cues are not abundant (e.g. a pedestrian is by its alone). Thus, we propose a simple overlap detector, which predicts the IoU between the $\m$ and the object overlapped with $\m$ the most. If the predicted IoU is less than a threshold $d_t$, which we empirically set to $0.3$, then NOH-Suppressor won't be triggered because of insufficient cues, and we will follow the design of Greedy-NMS ($s_i := 0$) or Soft-NMS ($s_i := s_if(\text{iou}(\m, b_i))$).

\textbf{NOH-Suppressor} If the cues are predicted to be sufficient, we will perform NOH-Suppression, which re-scores the $s_i$ by multiplying the probability of $b_i$ being a nearby object. In this way, when a neighboring box meets the attributes of being a nearby object, the suppression on it will be dynamically eased, whereas if it is very unlikely to be a nearby object, then we treat it as detecting the same object of $\m$, which should be degraded. We formally describe the difference between \nmsname{} and Greedy-NMS in Fig.~\ref{fig:algorithm}. As we only replace the re-scoring function with a Gaussian function runs at $O(1)$, we haven't introduced computational complexity into the NMS pipeline. In addition, since we leverage the mini 2-fc branch to predict both distribution parameters and density directly from RoI feature, the overhead is negligible (See Fig.~\ref{fig:architecture}).

In summary, the strategy we adopt can be described as follow:
\begin{equation}
    s_i=\left\{
        \begin{array}{ll}
            s_i,
            &\text{iou}(\m, b_i) < N_t 
            \\
            s_i \cdot f(\m, b_i, d_\m, p_\m),
            &\text{iou}(\m, b_i) \geq N_t
        \end{array}
        ,
        \right.
    \label{eq:our-nms}
\end{equation}

\begin{equation}
    f(\m, b_i, d_\m, p_\m)=\left\{
        \begin{array}{ll}
            P_{p_\m}(\m, b_i),
            &d_\m \geq d_t
            \\
            0,
            &d_\m < d_t
        \end{array}
        \right.
    \label{eq:our-nms-f}
\end{equation}

Note that, if the step function in Eq.~\ref{eq:adaptive-nms-f} is used as the PDF, then our \nmsname{} degenerates to Adaptive-NMS. However, the step function is rarely used for modeling the natural distributions because (1) it is not continuous, and (2) it is oversimplified. Thus, we propose \heatmapnameshort{} (Sec.~\ref{sec:heatmap}) to better capture the true nearby-objects distribution using the Gaussian distribution. 

\subsection{\heatmapname{} (\heatmapnameshort{})}
\label{sec:heatmap}
\heatmapnameshort{} is responsible for generating the nearby-objects distribution for each $\m$. We achieve this by hallucinating the locations and shapes of the nearby objects from the cues in region $\m$, and expressing the hallucination with a Gaussian distribution. We term this process as hallucination because different from proposal-based instance recognition, which predicts box coefficients from the proper RoI feature, our NOH could only rely on partially visible cues.

Essentially, based on the features extracted from region $\m$, multiple hallucination objects could be proposed. However, for simplicity, we only capture one nearby object which overlaps with $\m$ the most. We represent the hallucinated object with its relative center location, width, height with $\m$, denoted as $\mu_\m$. Since the hallucinated object is predicted by partially visible cues, the prediction is expected to be imprecise. Thus, we decay the nearby-objects likelihood with a Gaussian distribution which centers at $\mu_\m$ and spreads with a hyper-parameter $\sigma$.

With all the definition above, our NOH applies the following strategy:

\begin{equation}
    P_{\mu_\m}(\m, b_i) =
    \text{exp}(-\norm{b_{i|\m} - \mu_\m}^2 / 2\sigma^2)
    \label{eq:our-nms-g-1}
\end{equation}

\begin{equation}
    b_{i|\m} = \{
        \frac{x_{b_i}-x_\m}{w_\m},
        \frac{y_{b_i}-y_\m}{h_\m},
        \text{log}\frac{w_{b_i}}{w_\m},
        \text{log}\frac{h_{b_i}}{h_\m}
    \}
    \label{eq:our-nms-g-2}
\end{equation}

\begin{figure}[t]
\begin{center}
\includegraphics[width=1.0\linewidth]{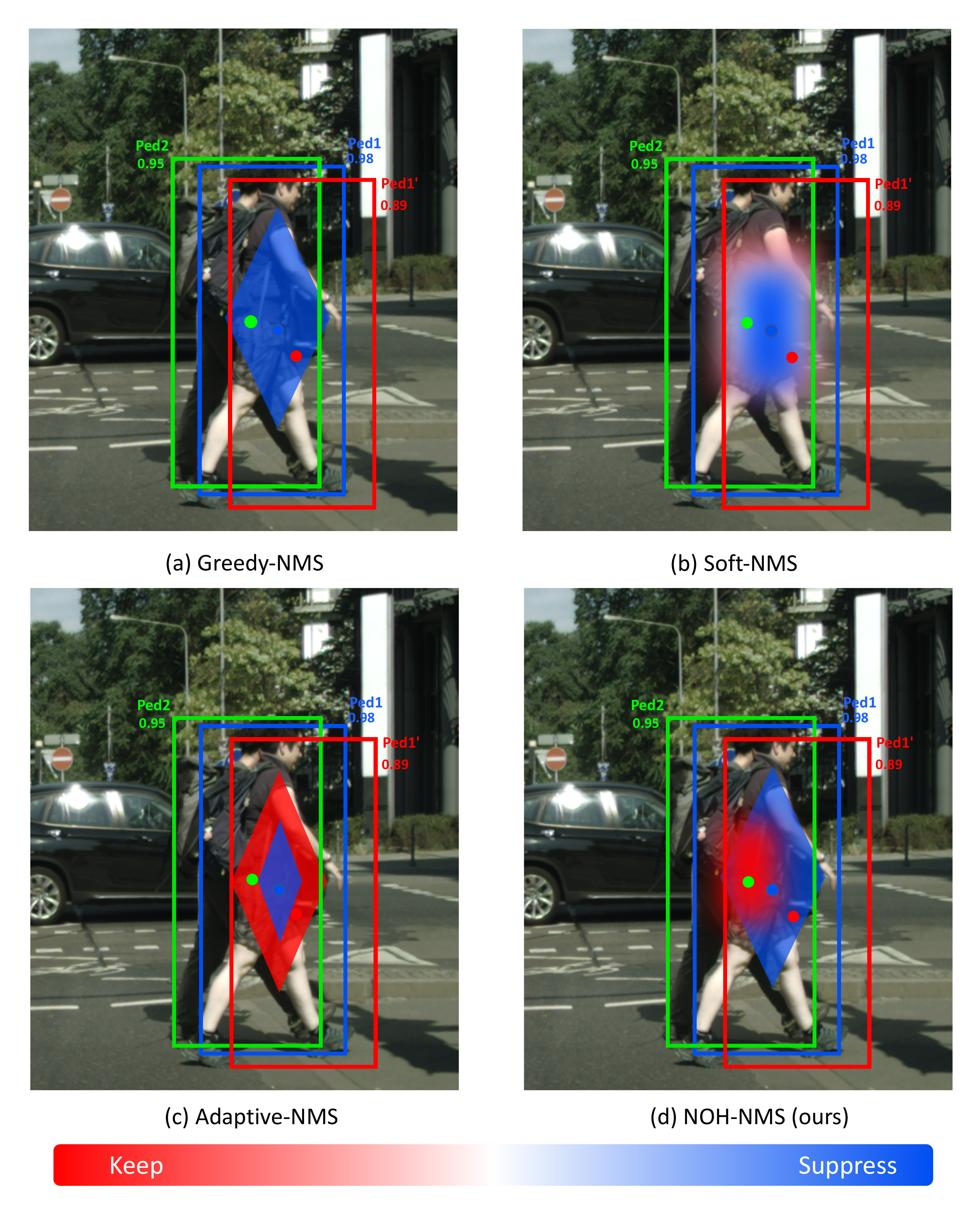}
\end{center}
\caption{\textbf{Visualization of the suppression degree} {\normalfont The suppression degree is a function of the relative center location and relative shape of two boxes, resulting in 4-d freedom. To visualize it in 2-d space, we unify the shapes of all the boxes so that each box can be represented by its center point. The detection score is attached to the corner of the box. The color map shows to what extent the detection with the maximum score (blue box) suppresses its surrounding inferior detections. For instance, the center point of the green box in (d) lies in the red area (keeping area), meaning it is very likely to survive the suppression, whereas the red box will be penalized harshly as its center sits in the blue area (suppressing area).}}
\vspace{-0.3cm}
\label{fig:nms}
\end{figure}


We implement NOH with a prediction head in parallel with the classification and regression head of Faster-RCNN. The training target of NOH is derived from the relative box coefficients of the most nearby object with $\m$, and we impose Smooth-L1 loss as the training loss. Note that, the Gaussian function is not represented during the training. However, we could convert the training target from the relative box coefficients into a Dirac delta function, and supervise it with KL Loss~\cite{kl-loss}. In this paper, we keep the training process simple, as we find it works up to the expectation, and stick with the Smooth-L1 loss.

\subsection{Comparisons with other NMS Strategies}
\label{sec:compare}
To better understand the difference among NMS strategies that we and other methods propose, we visualize the suppression effect of $\m$ on overlapped other detections in Fig.~\ref{fig:nms}. According to the figure, Greedy-NMS harshly eliminates the detections around $\m$, and Soft-NMS gradually adds \textit{keeping} area. Adaptive-NMS, on the other hand, adds a more harsh \textit{keeping} area, as the result of the usage of the step function, but the proportion of such area is adaptive to the pedestrian density. Note that when combining Soft-NMS and Adaptive-NMS together, the \textit{keeping} area will be both continuous and adaptive. However, all the aforementioned methods cannot shift the center of the \textit{keeping} area because they don't explicitly predict the distribution of the nearby pedestrians, whereas our method places the \textit{keeping} area more accurate thanks to the \heatmapnameshort{} module.

\section{Experiments}
\label{sec:experiments}
In this section, we first cover the datasets and metrics that we use for all the experiments (Sec.~\ref{sec:data}). We then reveal our implementation details in Sec.~\ref{sec:impl} and show quantitative results of \nmsname{} compared to various NMS methods (Sec.~\ref{sec:results}). We also conduct sensitivity analysis (Sec.~\ref{sec:sensitivity}) to prove the robustness of our method. Qualitative results are also prepared in Sec.~\ref{sec:qualitative} for better visualization.

\subsection{Datasets and Metrics}
\label{sec:data}
\textbf{CityPersons} CityPersons~\cite{citypersons} is a currently wildly used benchmark dataset in the pedestrian detection task. Based on the $5000$ images in the Cityscapes~\cite{cityscapes} dataset, CityPersons creates more fine-grained bounding box annotations which dedicate to pedestrian detection. In total, CityPersons covers $\sim35$k person and $\sim13k$ ignore region (fake humans like statues) annotations. In addition, CityPersons aims at including persons with heavy occlusion and small scale, yielding an average density of $\sim7$ persons per image. 

\textbf{CrowdHuman} CrowdHuman~\cite{crowdhuman} was released more recently, which further emphasizes the crowd issue. It contains $15,000$ images, with $\sim340$k person and $\sim99$k ignore region annotations. The person density is significantly higher than CityPersons and reaches $\sim22.6$ persons per image with $2.4$ pairwise overlapping instances (IoU larger than $0.5$).

\textbf{Evaluation metrics} We follow the evaluation metrics used in CityPersons and CrowdHuman, denotes as MR\textsuperscript{-2}, AP, and Recall:
\begin{itemize}
    \item MR\textsuperscript{-2}, or log-average Miss Rate on False Positive Per Image (FPPI) in $[10^{-2}, 10^0]$, is commonly used to evaluate detectors whose applications have an upper limit on the acceptable FPPI rate independent of object density. Thus, MR\textsuperscript{-2} is particularly sensitive to false positives. 
    \item Average Precision (AP) is the most popular metric in generic object detection, which summarizes the precision-recall curve of the detection results. In the following experiments, we follow the AP metric in PASCAL VOC~\cite{pascal}, where a prediction is positive if IoU $\geq 0.5$.
    \item Recall is short for the maximum recall given a fixed number of detections. As both Soft-NMS, Adaptive-NMS, and \nmsname{} aim at recalling the mis-eliminated true positives, as shown in Fig~.\ref{fig:nms}, this metric reflects the effectiveness of this intention. For fair comparisons, we set the allowed number of detections to be $100$ for all NMS methods.
\end{itemize}


\begin{table*}[t]
\begin{center}
\begin{tabular}{cccccccc}
\hline
Methods & Extra Anno.   & Backbone & Scale & Reasonable & Bare & Partial & Heavy  \\
\hline
OR-CNN~\cite{orcnn} & \checkmark    & VGG-16 & $\times 1.3$& 11.0 & \textbf{5.9} & \textbf{13.7} & 51.3\\
MGAN~\cite{mgan} & \checkmark    & VGG-16 &$\times 1.3$& 10.5 & - &- & \textbf{47.2}\\
JointDet~\cite{jointdet}& \checkmark    & ResNet-50 &$\times 1.3$&  \textbf{10.2} & -& - & -\\
\hline
 TLL (MRF)~\cite{tll} && ResNet-50 & - & 14.4 & - & - & - \\
Adapted Faster RCNN~\cite{citypersons} && VGG-16 & $\times 1.3$ & 13.0 & - & - & -\\
ALFNet~\cite{alf} && VGG-16 & $\times 1$& 12.0 & 8.4 & 11.4 & \textbf{51.9}\\
RepLoss~\cite{repulsion-loss} && ResNet-50 & $\times 1.3$& 11.6 & 7.0 & 14.8 & 55.3 \\
Adaptive-NMS w/ AggLoss~\cite{adaptive-nms} &&  VGG-16 & $\times 1.3$& \textbf{10.8} & \textbf{6.2} & 11.4 & 54.0\\
Our baseline && ResNet-50 &$\times 1.3$& 11.9 & 7.4 & 12.3 & 53.0\\
\nmsname{} && ResNet-50 & $\times 1.3$& \textbf{10.8} & 6.6 & \textbf{11.2} & 53.0\\
\end{tabular}
\end{center}
\caption{Performance on the CityPersons validation set. {\normalfont \mr{} is used as the metric (lower is better). Scale is short for input scale.}}
\label{tab:citypersons}
\end{table*}

\subsection{Implementation Details}
\label{sec:impl}
For all the experiments, we adapt the Faster-RCNN~\cite{faster-rcnn} with FPN~\cite{fpn} as our baseline and build various NMS methods upon the same baseline for fair comparisons. In specific, we choose the standard ResNet-50~\cite{resnet} as the backbone and replace the ROIPooling operation in the original Faster-RCNN with the RoIAlign~\cite{mask-rcnn}. We also change the aspect ratios of the anchors to $H/W=\{1, 2, 3\}$ for CrowdHuman and $H/W=\{2.44\}$ for CityPersons, as the original anchor settings are optimized towards COCO~\cite{coco}. Following the choice of input size in~\cite{citypersons} and~\cite{crowdhuman}, we enlarge the input height and width of CityPersons by $1.3$ times and resize the input of CrowdHuman so that the shorter edge of input equals to 800 pixels while keeping the longer edge no longer than 1,400 pixels.

During training, we randomly initialize all the parameters of the model by Kaiming initialization~\cite{kaiming-init}, except the ResNet-50 backbone, whose initial parameters are loaded from ImageNet~\cite{imagenet} pre-train. We use SGD with $0.9$ momentum and $0.0001$ weight decay as the optimizer and train the model with $5,600$ and $28,125$ iterations in total for CityPersons and CrowdHuman respectively. The initial learning rate is $0.02 (0.04)$ and decreases by a factor of $10$ after $3,400 (18,750)$ and $4,600 (24,375)$ iterations for CityPersons (CrowdHuman). The batch size is set to be $16$ for both datasets. Note that we train on 8 GPUs without Synchronized BN.

For CityPersons, a sample will be assigned as positive if its IoU with ground-truth is greater than $0.7$, and as negative if the IoU is less than $0.5$, otherwise the sample will be ignored and won't contribute to the loss. For CrowdHuman, samples with IoU greater than $0.5$ qualify as the positive and otherwise are considered as negative. In addition, we clip the ground-truth bounding boxes at the image boundary for CityPersons, while don't apply this operation in CrowdHuman.

During inference, we set the NMS IoU threshold to $0.5$ for all NMS methods and allow at most $100$ detections per image. We also follow the same input resizing operation as mentioned in the training stage.



\begin{table}[t]
\begin{center}
\begin{tabular}{ccccc}
\hline
Methods & Backbone  & AP & Recall & \mr \\
\hline
Repulsion Loss~\cite{repulsion-loss} & R50 & - & - & 45.7 \\
JointDet*~\cite{jointdet} & R50 & -& - & 46.5 \\
Baseline in~\cite{adaptive-nms} & R50 & 83.0 & 90.6 & 52.4\\
Adaptive-NMS~\cite{adaptive-nms} & R50 & 84.7 & 91.3 & 49.7\\
Our Baseline & R50 & 85.1 & 87.8 & 44.7\\
\nmsname{} & R50 & \textbf{89.0} & \textbf{92.9} & \textbf{43.9}\\

\end{tabular}
\end{center}
\caption{Performance on the CrowdHuman validation set. {\normalfont R50 denotes ResNet-50. * marks the methods which leverage extra annotations (e.g. head box) during training.}}
\vspace{-0.5cm}
\label{tab:crowdhuman}
\end{table}


\begin{table}[t]
\begin{center}
\begin{tabular}{ccccc}
\hline
Methods & $N_t$  & AP & Recall & $MR^{-2}$ \\
\hline
Greedy-NMS & 0.5 & 85.1 & 87.8 & 44.7\\
Soft-NMS~\cite{soft-nms} & 0.5 & 86.4 & 90.6 & 44.6 \\
Adaptive-NMS~\cite{adaptive-nms} & 0.5 & 87.1 & 89.2 & 45.0\\
\nmsname{} & 0.5 & \textbf{89.0} & \textbf{92.9} & \textbf{43.9}\\

\end{tabular}
\end{center}
\caption{Comparison of different NMS methods on the CrowdHuman validation set. {\normalfont All the methods are implemented by us, and for fair comparisons, we show the best results from multiple runs.}}
\vspace{-0.5cm}
\label{tab:ablation_nms}
\end{table}

\subsection{Results}
\label{sec:results}

\textbf{CityPersons} We report the results of \nmsname{} and other state-of-the-art pedestrian detectors on CityPersons validation set in Tab.~\ref{tab:citypersons}. In particular, according to the level of occlusion, the CityPersons has four splits, namely Bare, Partial, Reasonable, and Heavy, whose ratios of visible parts are $[0.9, 1], [0.65, 0.9], [0.65, 1], [0.2, 0.65]$. Within the group of the methods which don't use extra annotation, \nmsname{} achieves the best performance on Reasonable, which is the most valued, and Partial splits. Moreover, our performance is comparable to that of the methods using additional annotations (e.g. head bounding boxes, visible bounding boxes).

\textbf{CrowdHuman} 
Tab.~\ref{tab:crowdhuman} shows the performance on CrowdHuman validation set. To have a comprehensive evaluation, three evaluation metrics are chosen to evaluate our method, which are AP, Recall, and \mr{}. We re-implement a strong FPN~\cite{fpn} baseline. Our baseline achieves $85.1\%$ AP, $87.8\%$ Recall and $44.7\%$ \mr{}, which outperforms the baseline in Adaptive-NMS~\cite{adaptive-nms} by $0.4\%$ AP and $5.0 \%$ \mr{}. Although compared to our strong Greedy-NMS baseline, \nmsname{} still significantly improves the AP, Recall, and \mr{} by $3.9\%$, $5.1\%$, and $0.8\%$. Moreover, compared to other state-of-the-art methods, superior performance demonstrates the effectiveness of our method.

To better demonstrate that our performance gain is not from the strong baseline, and show more clearly the advantage of \nmsname{} compared with its counterparts, we re-implement Soft-NMS and Adaptive-NMS on our strong baseline. The results are shown in Tab.~\ref{tab:ablation_nms}. According to the results, \nmsname{} still delivers the best performance across all the evaluation metrics.

\begin{figure*}
    \includegraphics[width=0.32\textwidth]{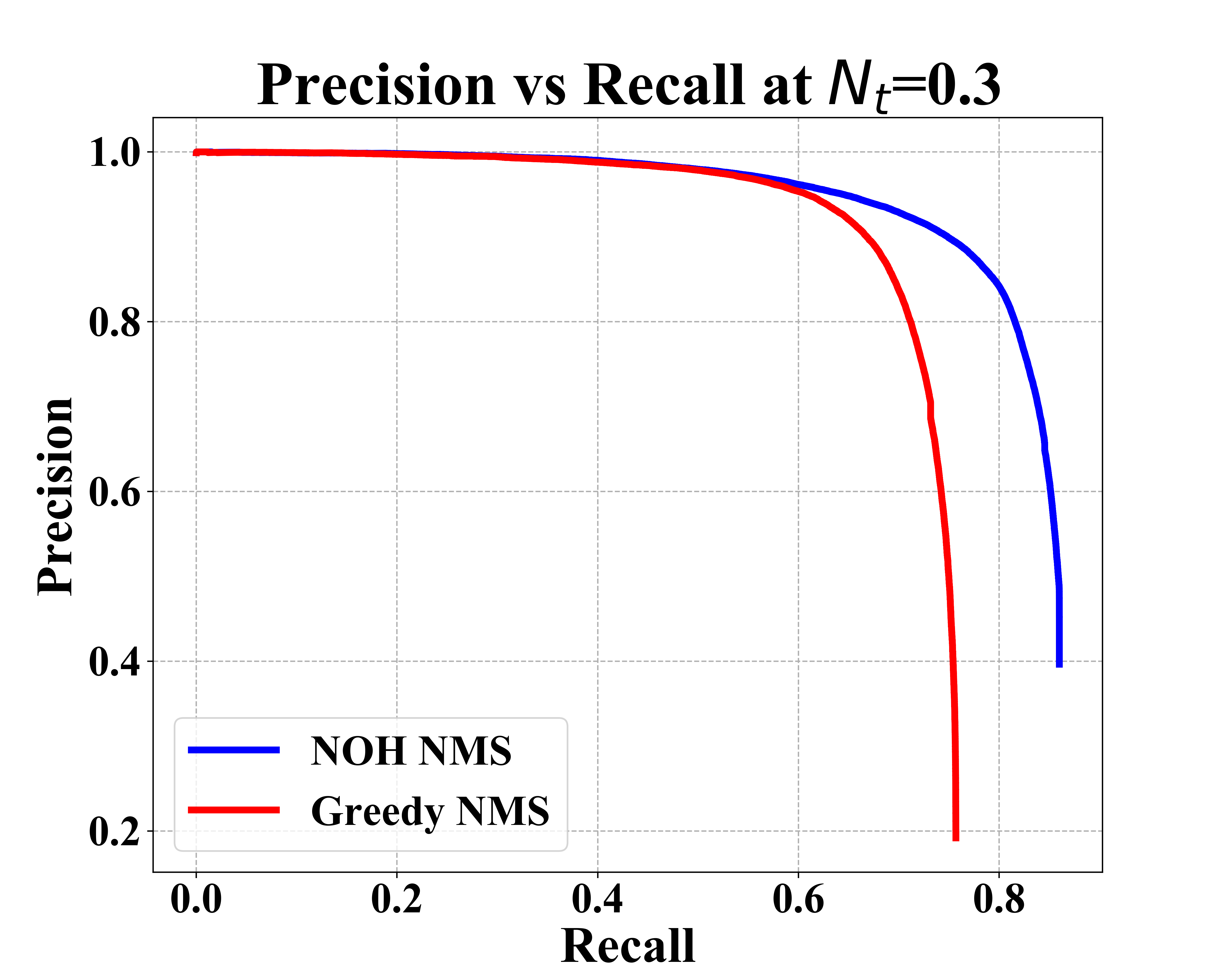}
    \includegraphics[width=0.32\textwidth]{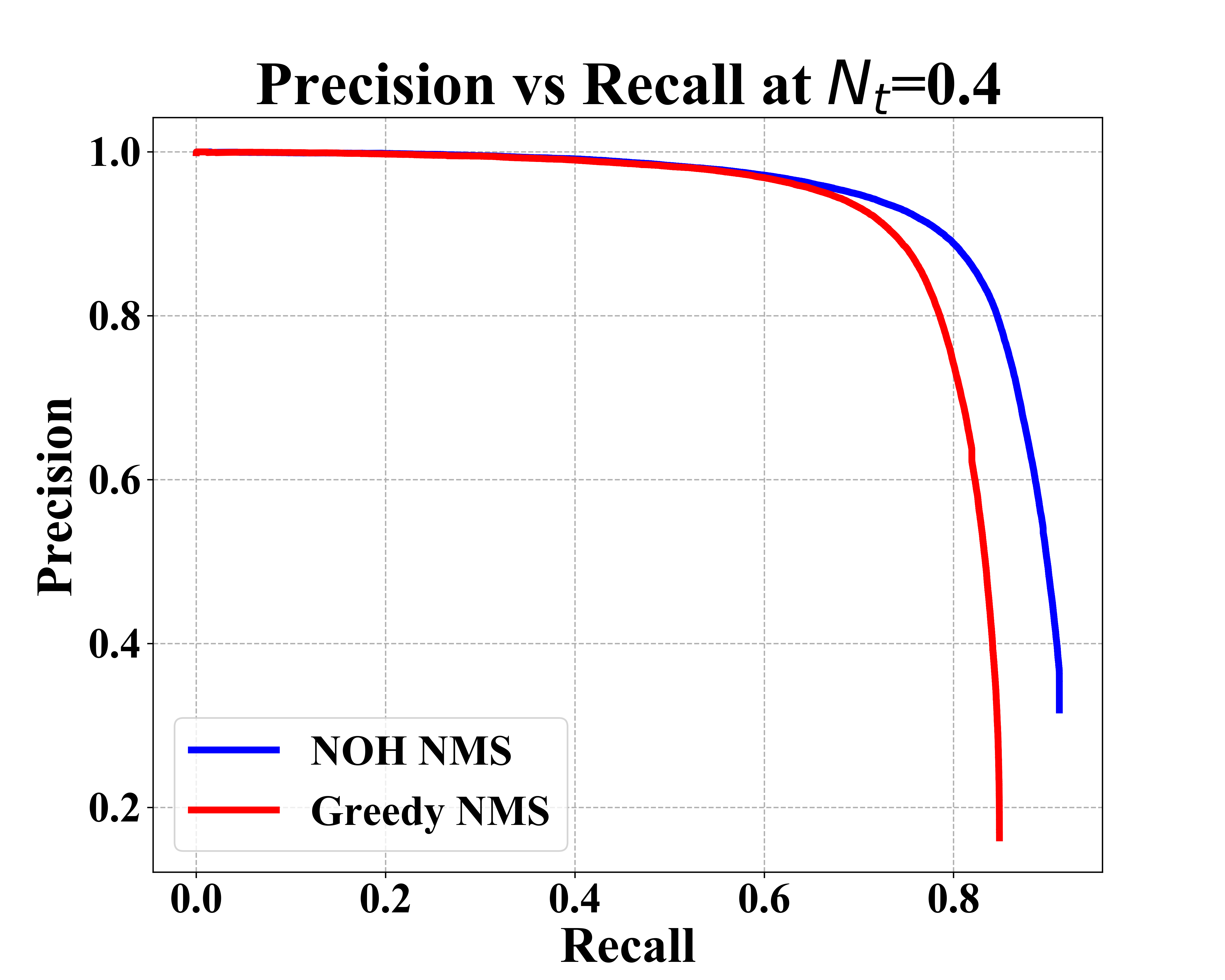}
    \includegraphics[width=0.32\textwidth]{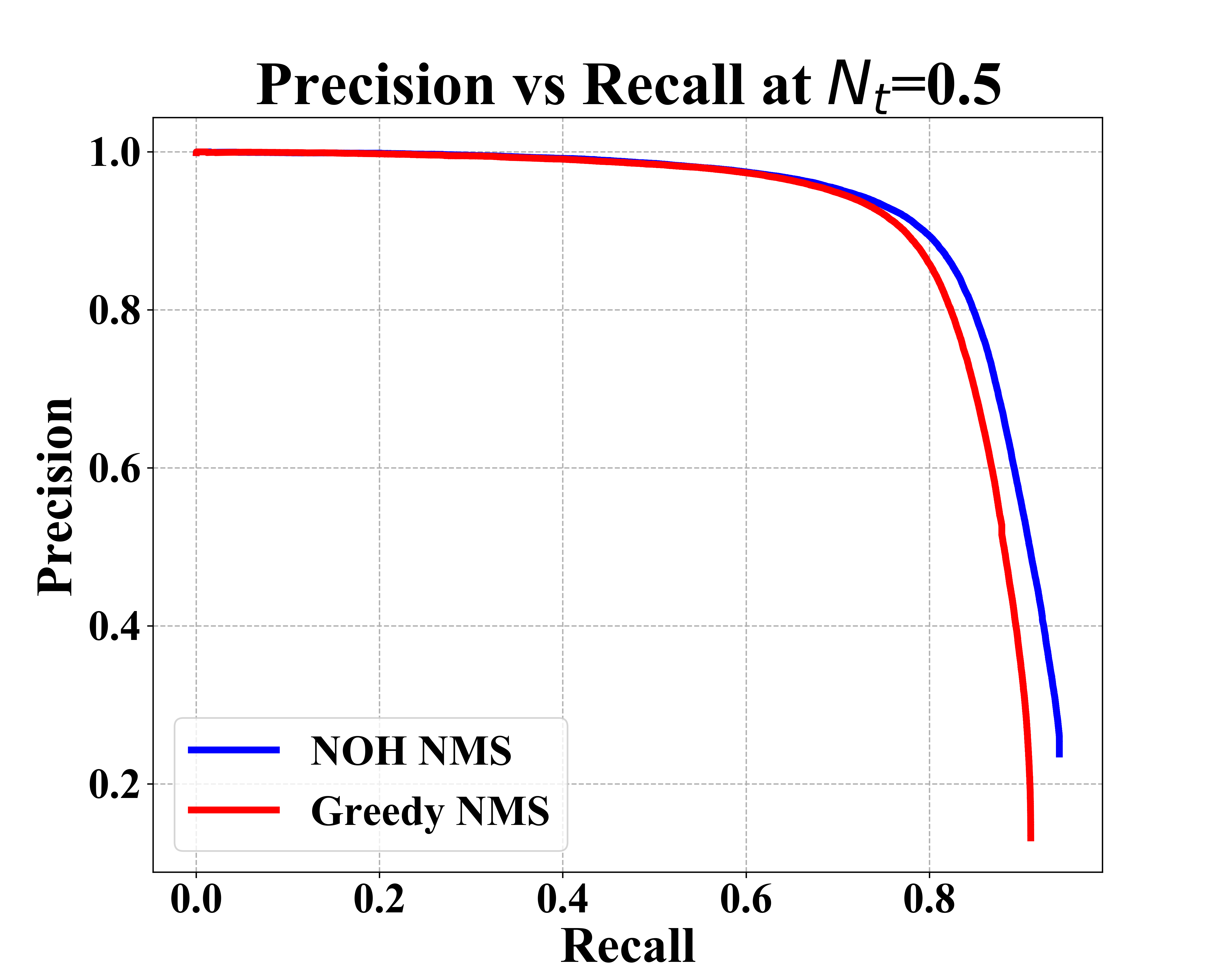}
    \vspace{-0.2cm}
    \caption{Precision vs. Recall at multiple NMS IoU thresholds $\mathbf{N_t}$ {\normalfont Experiments are conducted on the CrowdHuman validation set and all the NMS methods are implemented by us based on the same baseline.}}
    \label{fig:pr}
\end{figure*}
\begin{figure*}
    \vspace{-0.3cm}
    \includegraphics[width=0.48\textwidth]{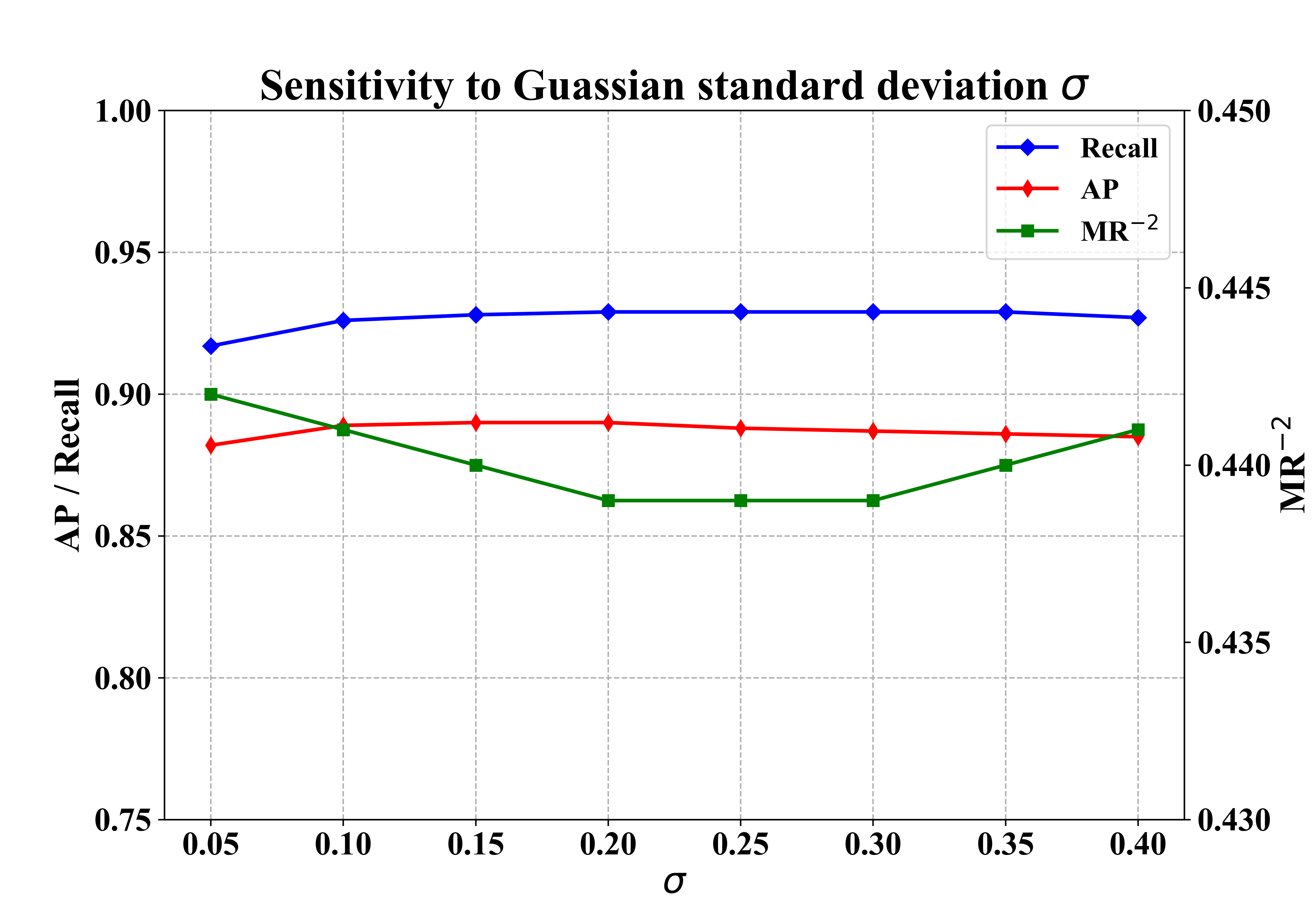}
    \includegraphics[width=0.48\textwidth]{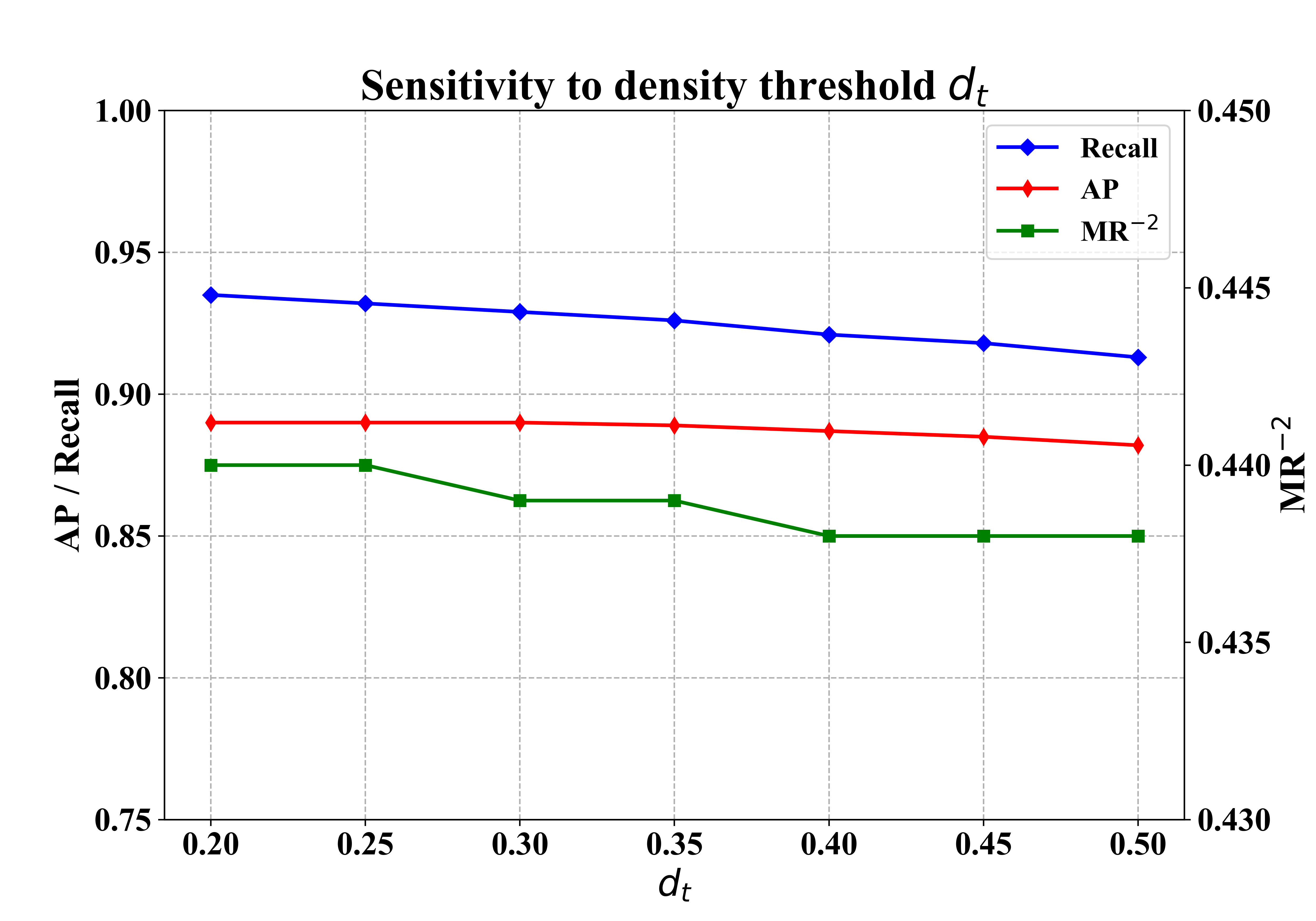}
    \vspace{-0.3cm}
    \caption{Sensitivity to hyper-parameters {\normalfont We show the effect of the different choices of $\sigma$ and $d_t$ on \nmsname{}. All the experiments are done on the CrowdHuman validation set.} }
    \label{fig:sigma}
\end{figure*}

\subsection{Sensitivity Analysis}
\label{sec:sensitivity}
Although \nmsname{} introduces two more hyper-parameters (density threshold $d_t$ and Gaussian standard deviation $\sigma$) than the other NMS methods, as we analyze later, it is not only robust to the choice of $d_t$ and $\sigma$, but also less sensitive to the common hyper-parameter $N_t$ than other NMS.

\textbf{IoU threshold} As shown in Fig.~\ref{fig:pr}, we plot the precision vs. recall curves on various NMS IoU thresholds for both Greedy-NMS and \nmsname{}. We conclude two points from the figure. (1) Even though both methods degrade with sub-optimal IoU threshold hyper-parameter, \nmsname{} is less sensitive as it outperforms the Greedy-NMS in all recall levels across all the choice of $N_t$. (2) Simply flexing the IoU threshold for Greedy-NMS does recall more true positives but also introduces even more false positives that overwhelm the overall performance.

\textbf{Density threshold} As one of the additional hyper-parameters we introduce, the density threshold $d_t$ determines what it takes to be considered as having abundant cues to support the existence of other nearby pedestrians. As shown in Fig.~\ref{fig:sigma}, the performance for AP, recall, and MR\textsuperscript{-2} jitters slightly with a wild range of $d_t$ (from $0.2$ to $0.5$ with an interval of $0.05$), which proves the robustness of \nmsname{}.

\textbf{Gaussian standard deviation} $\sigma$ controls the spread of the Gaussian distribution we use in \heatmapnameshort{}. Even though we empirically set it to $0.2$ in our previous experiments, it is proven to be not very sensitive as illustrated in Fig.~\ref{fig:sigma}. Note that, if using KL Loss during training, the $\sigma$ can be trained end-to-end, and will no longer be a hyper-parameter. However, we leave this as future work since it is not the focus of this paper.

\subsection{Qualitative Results}
\label{sec:qualitative}
\begin{figure*}[t]
\begin{center}
\includegraphics[width=1.0\linewidth]{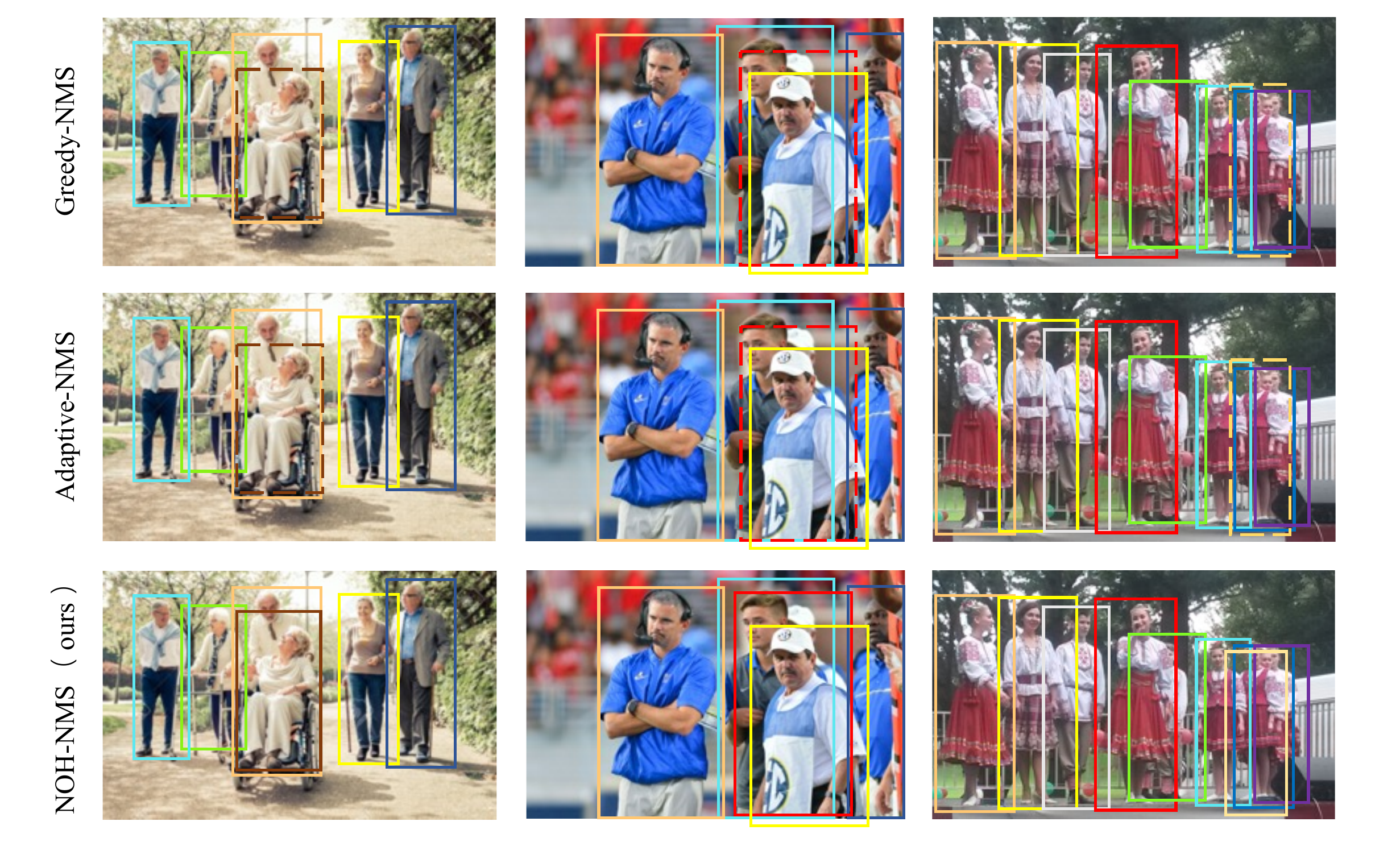}
\end{center}
\vspace{-0.8cm}
\caption{\textbf{Qualitative results} {\normalfont Evaluation results on the CrowdHuman validation set. The NMS IoU threshold is set to $0.5$ for all the methods. The dotted boxes show the missing detections.}}
\vspace{-0.2cm}
\label{fig:experiments}
\end{figure*}

\begin{figure*}[t]
\begin{center}
\includegraphics[width=0.9\linewidth]{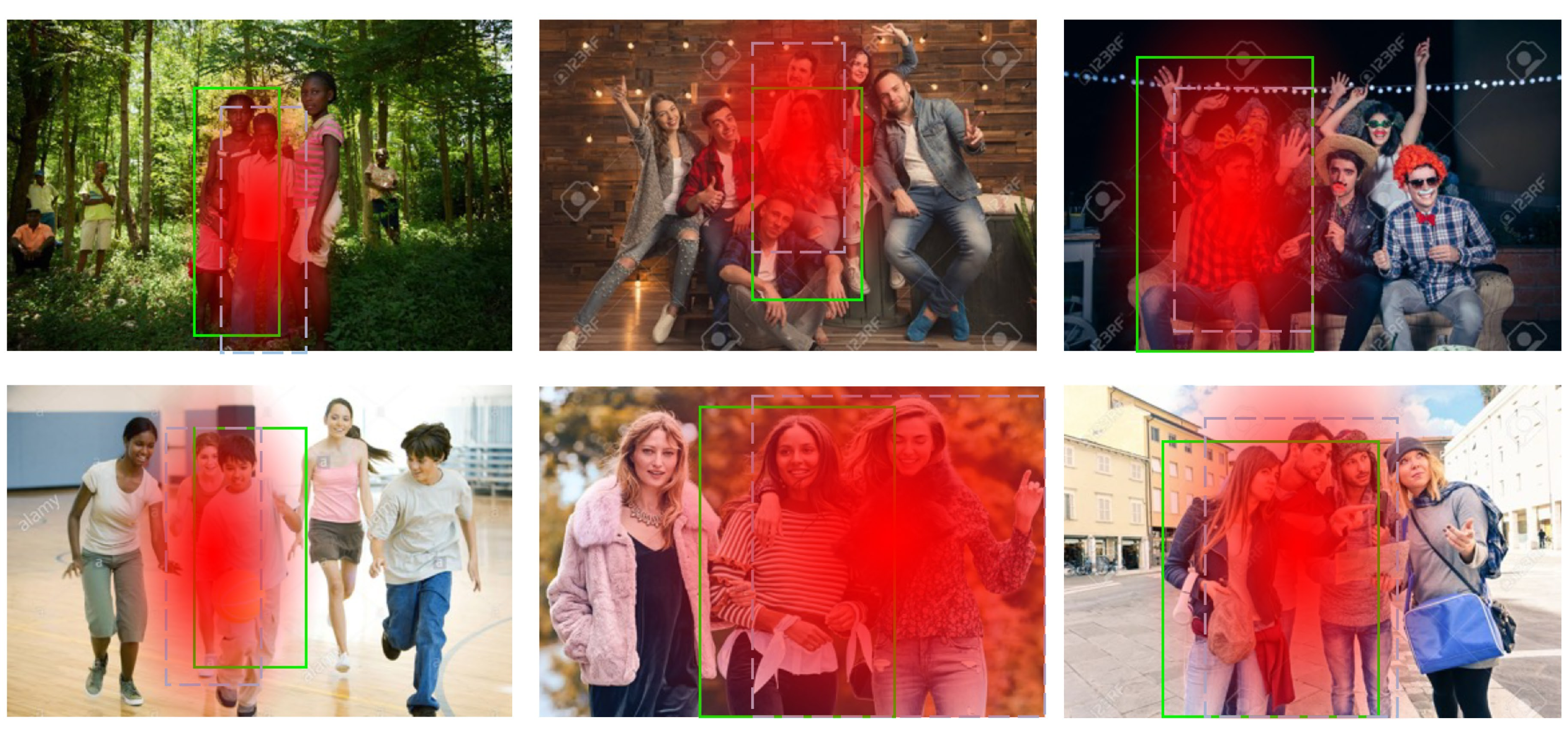}
\end{center}
\vspace{-0.3cm}
\caption{\textbf{The visualization of the nearby objects hallucination results} {\normalfont \heatmapnameshort{} models the distribution of nearby objects with a 4-d Gaussian whose mean $\mu_\m$ represents the expectation of the location and shape of the nearest object (shown in the dotted blue box). The variance of the 2-d transition of the center points is illustrated in red (we don't show the shape variance). The green boxes show the prediction for $\m$.}}
\label{fig:gaussian_visualize}
\end{figure*}

Qualitative results are given in two aspects: (1) detections visualization compared with Greedy-NMS and Adaptive-NMS (Fig.~\ref{fig:experiments}); (2) illustration of the effectiveness of the nearby objects hallucination (Fig.~\ref{fig:gaussian_visualize}).

As shown in Fig.~\ref{fig:experiments}, our \nmsname{} successfully recalls the highly overlapped detections that other methods fail to do so. Moreover, in Fig.~\ref{fig:gaussian_visualize}, the \heatmapname{} works as expect, pinpointing the nearby persons with a reasonable Gaussian distribution, which contributes significantly to helping \nmsname{} ease the suppression on the highly overlapped areas.

\section{Conclusion}

In this paper, we present a novel \nmsname{} algorithm that improves the performance of pedestrian detection by taking into account the distribution of nearby objects. As the core part of our algorithm, \heatmapname{} learns to predict the Gaussian distribution of nearby objects from only full-body box annotations and introduces marginal overhead. Comprehensive experiments and analyses are done on CityPersons \cite{citypersons} and CrowdHuman \cite{crowdhuman} to show the strength of \nmsname{}.

\newpage
\bibliographystyle{ACM-Reference-Format}
\bibliography{Noh-NMS}


\begin{thebibliography}{37}


\ifx \showCODEN    \undefined \def \showCODEN     #1{\unskip}     \fi
\ifx \showDOI      \undefined \def \showDOI       #1{#1}\fi
\ifx \showISBNx    \undefined \def \showISBNx     #1{\unskip}     \fi
\ifx \showISBNxiii \undefined \def \showISBNxiii  #1{\unskip}     \fi
\ifx \showISSN     \undefined \def \showISSN      #1{\unskip}     \fi
\ifx \showLCCN     \undefined \def \showLCCN      #1{\unskip}     \fi
\ifx \shownote     \undefined \def \shownote      #1{#1}          \fi
\ifx \showarticletitle \undefined \def \showarticletitle #1{#1}   \fi
\ifx \showURL      \undefined \def \showURL       {\relax}        \fi
\providecommand\bibfield[2]{#2}
\providecommand\bibinfo[2]{#2}
\providecommand\natexlab[1]{#1}
\providecommand\showeprint[2][]{arXiv:#2}

\bibitem[\protect\citeauthoryear{Bodla, Singh, Chellappa, and Davis}{Bodla
  et~al\mbox{.}}{2017}]%
        {soft-nms}
\bibfield{author}{\bibinfo{person}{Navaneeth Bodla}, \bibinfo{person}{Bharat
  Singh}, \bibinfo{person}{Rama Chellappa}, {and} \bibinfo{person}{Larry~S
  Davis}.} \bibinfo{year}{2017}\natexlab{}.
\newblock \showarticletitle{Soft-NMS--improving object detection with one line
  of code}. In \bibinfo{booktitle}{\emph{Proceedings of the IEEE international
  conference on computer vision}}. \bibinfo{pages}{5561--5569}.
\newblock


\bibitem[\protect\citeauthoryear{Cai and Vasconcelos}{Cai and
  Vasconcelos}{2018}]%
        {cascade-rcnn}
\bibfield{author}{\bibinfo{person}{Zhaowei Cai} {and} \bibinfo{person}{Nuno
  Vasconcelos}.} \bibinfo{year}{2018}\natexlab{}.
\newblock \showarticletitle{Cascade r-cnn: Delving into high quality object
  detection}. In \bibinfo{booktitle}{\emph{Proceedings of the IEEE conference
  on computer vision and pattern recognition}}. \bibinfo{pages}{6154--6162}.
\newblock


\bibitem[\protect\citeauthoryear{Chi, Zhang, Xing, Lei, Li, and Zou}{Chi
  et~al\mbox{.}}{2019}]%
        {jointdet}
\bibfield{author}{\bibinfo{person}{Cheng Chi}, \bibinfo{person}{Shifeng Zhang},
  \bibinfo{person}{Junliang Xing}, \bibinfo{person}{Zhen Lei},
  \bibinfo{person}{Stan~Z Li}, {and} \bibinfo{person}{Xudong Zou}.}
  \bibinfo{year}{2019}\natexlab{}.
\newblock \showarticletitle{Relational Learning for Joint Head and Human
  Detection}.
\newblock \bibinfo{journal}{\emph{arXiv preprint arXiv:1909.10674}}
  (\bibinfo{year}{2019}).
\newblock


\bibitem[\protect\citeauthoryear{Cordts, Omran, Ramos, Rehfeld, Enzweiler,
  Benenson, Franke, Roth, and Schiele}{Cordts et~al\mbox{.}}{2016}]%
        {cityscapes}
\bibfield{author}{\bibinfo{person}{Marius Cordts}, \bibinfo{person}{Mohamed
  Omran}, \bibinfo{person}{Sebastian Ramos}, \bibinfo{person}{Timo Rehfeld},
  \bibinfo{person}{Markus Enzweiler}, \bibinfo{person}{Rodrigo Benenson},
  \bibinfo{person}{Uwe Franke}, \bibinfo{person}{Stefan Roth}, {and}
  \bibinfo{person}{Bernt Schiele}.} \bibinfo{year}{2016}\natexlab{}.
\newblock \showarticletitle{The cityscapes dataset for semantic urban scene
  understanding}. In \bibinfo{booktitle}{\emph{Proceedings of the IEEE
  conference on computer vision and pattern recognition}}.
  \bibinfo{pages}{3213--3223}.
\newblock


\bibitem[\protect\citeauthoryear{Dai, Li, He, and Sun}{Dai
  et~al\mbox{.}}{2016}]%
        {rfcn}
\bibfield{author}{\bibinfo{person}{Jifeng Dai}, \bibinfo{person}{Yi Li},
  \bibinfo{person}{Kaiming He}, {and} \bibinfo{person}{Jian Sun}.}
  \bibinfo{year}{2016}\natexlab{}.
\newblock \showarticletitle{R-fcn: Object detection via region-based fully
  convolutional networks}. In \bibinfo{booktitle}{\emph{Advances in neural
  information processing systems}}. \bibinfo{pages}{379--387}.
\newblock


\bibitem[\protect\citeauthoryear{Dai, Qi, Xiong, Li, Zhang, Hu, and Wei}{Dai
  et~al\mbox{.}}{2017}]%
        {deformable}
\bibfield{author}{\bibinfo{person}{Jifeng Dai}, \bibinfo{person}{Haozhi Qi},
  \bibinfo{person}{Yuwen Xiong}, \bibinfo{person}{Yi Li},
  \bibinfo{person}{Guodong Zhang}, \bibinfo{person}{Han Hu}, {and}
  \bibinfo{person}{Yichen Wei}.} \bibinfo{year}{2017}\natexlab{}.
\newblock \showarticletitle{Deformable convolutional networks}. In
  \bibinfo{booktitle}{\emph{Proceedings of the IEEE international conference on
  computer vision}}. \bibinfo{pages}{764--773}.
\newblock


\bibitem[\protect\citeauthoryear{Everingham, Eslami, Van~Gool, Williams, Winn,
  and Zisserman}{Everingham et~al\mbox{.}}{2015}]%
        {pascal}
\bibfield{author}{\bibinfo{person}{Mark Everingham}, \bibinfo{person}{SM~Ali
  Eslami}, \bibinfo{person}{Luc Van~Gool}, \bibinfo{person}{Christopher~KI
  Williams}, \bibinfo{person}{John Winn}, {and} \bibinfo{person}{Andrew
  Zisserman}.} \bibinfo{year}{2015}\natexlab{}.
\newblock \showarticletitle{The pascal visual object classes challenge: A
  retrospective}.
\newblock \bibinfo{journal}{\emph{International journal of computer vision}}
  \bibinfo{volume}{111}, \bibinfo{number}{1} (\bibinfo{year}{2015}),
  \bibinfo{pages}{98--136}.
\newblock


\bibitem[\protect\citeauthoryear{Fu, Liu, Ranga, Tyagi, and Berg}{Fu
  et~al\mbox{.}}{2017}]%
        {dssd}
\bibfield{author}{\bibinfo{person}{Cheng-Yang Fu}, \bibinfo{person}{Wei Liu},
  \bibinfo{person}{Ananth Ranga}, \bibinfo{person}{Ambrish Tyagi}, {and}
  \bibinfo{person}{Alexander~C Berg}.} \bibinfo{year}{2017}\natexlab{}.
\newblock \showarticletitle{Dssd: Deconvolutional single shot detector}.
\newblock \bibinfo{journal}{\emph{arXiv preprint arXiv:1701.06659}}
  (\bibinfo{year}{2017}).
\newblock


\bibitem[\protect\citeauthoryear{Girshick}{Girshick}{2015}]%
        {fast-rcnn}
\bibfield{author}{\bibinfo{person}{Ross Girshick}.}
  \bibinfo{year}{2015}\natexlab{}.
\newblock \showarticletitle{Fast r-cnn}. In
  \bibinfo{booktitle}{\emph{Proceedings of the IEEE international conference on
  computer vision}}. \bibinfo{pages}{1440--1448}.
\newblock


\bibitem[\protect\citeauthoryear{Girshick, Donahue, Darrell, and
  Malik}{Girshick et~al\mbox{.}}{2014}]%
        {rcnn}
\bibfield{author}{\bibinfo{person}{Ross Girshick}, \bibinfo{person}{Jeff
  Donahue}, \bibinfo{person}{Trevor Darrell}, {and} \bibinfo{person}{Jitendra
  Malik}.} \bibinfo{year}{2014}\natexlab{}.
\newblock \showarticletitle{Rich feature hierarchies for accurate object
  detection and semantic segmentation}. In
  \bibinfo{booktitle}{\emph{Proceedings of the IEEE conference on computer
  vision and pattern recognition}}. \bibinfo{pages}{580--587}.
\newblock


\bibitem[\protect\citeauthoryear{He, Gkioxari, Doll{\'a}r, and Girshick}{He
  et~al\mbox{.}}{2017}]%
        {mask-rcnn}
\bibfield{author}{\bibinfo{person}{Kaiming He}, \bibinfo{person}{Georgia
  Gkioxari}, \bibinfo{person}{Piotr Doll{\'a}r}, {and} \bibinfo{person}{Ross
  Girshick}.} \bibinfo{year}{2017}\natexlab{}.
\newblock \showarticletitle{Mask r-cnn}. In
  \bibinfo{booktitle}{\emph{Proceedings of the IEEE international conference on
  computer vision}}. \bibinfo{pages}{2961--2969}.
\newblock


\bibitem[\protect\citeauthoryear{He, Zhang, Ren, and Sun}{He
  et~al\mbox{.}}{2015}]%
        {kaiming-init}
\bibfield{author}{\bibinfo{person}{Kaiming He}, \bibinfo{person}{Xiangyu
  Zhang}, \bibinfo{person}{Shaoqing Ren}, {and} \bibinfo{person}{Jian Sun}.}
  \bibinfo{year}{2015}\natexlab{}.
\newblock \showarticletitle{Delving deep into rectifiers: Surpassing
  human-level performance on imagenet classification}. In
  \bibinfo{booktitle}{\emph{Proceedings of the IEEE international conference on
  computer vision}}. \bibinfo{pages}{1026--1034}.
\newblock


\bibitem[\protect\citeauthoryear{He, Zhang, Ren, and Sun}{He
  et~al\mbox{.}}{2016}]%
        {resnet}
\bibfield{author}{\bibinfo{person}{Kaiming He}, \bibinfo{person}{Xiangyu
  Zhang}, \bibinfo{person}{Shaoqing Ren}, {and} \bibinfo{person}{Jian Sun}.}
  \bibinfo{year}{2016}\natexlab{}.
\newblock \showarticletitle{Deep residual learning for image recognition}. In
  \bibinfo{booktitle}{\emph{Proceedings of the IEEE conference on computer
  vision and pattern recognition}}. \bibinfo{pages}{770--778}.
\newblock


\bibitem[\protect\citeauthoryear{He, Zhu, Wang, Savvides, and Zhang}{He
  et~al\mbox{.}}{2019}]%
        {kl-loss}
\bibfield{author}{\bibinfo{person}{Yihui He}, \bibinfo{person}{Chenchen Zhu},
  \bibinfo{person}{Jianren Wang}, \bibinfo{person}{Marios Savvides}, {and}
  \bibinfo{person}{Xiangyu Zhang}.} \bibinfo{year}{2019}\natexlab{}.
\newblock \showarticletitle{Bounding box regression with uncertainty for
  accurate object detection}. In \bibinfo{booktitle}{\emph{Proceedings of the
  IEEE Conference on Computer Vision and Pattern Recognition}}.
  \bibinfo{pages}{2888--2897}.
\newblock


\bibitem[\protect\citeauthoryear{Lin, Doll{\'a}r, Girshick, He, Hariharan, and
  Belongie}{Lin et~al\mbox{.}}{2017a}]%
        {fpn}
\bibfield{author}{\bibinfo{person}{Tsung-Yi Lin}, \bibinfo{person}{Piotr
  Doll{\'a}r}, \bibinfo{person}{Ross Girshick}, \bibinfo{person}{Kaiming He},
  \bibinfo{person}{Bharath Hariharan}, {and} \bibinfo{person}{Serge Belongie}.}
  \bibinfo{year}{2017}\natexlab{a}.
\newblock \showarticletitle{Feature pyramid networks for object detection}. In
  \bibinfo{booktitle}{\emph{Proceedings of the IEEE conference on computer
  vision and pattern recognition}}. \bibinfo{pages}{2117--2125}.
\newblock


\bibitem[\protect\citeauthoryear{Lin, Goyal, Girshick, He, and Doll{\'a}r}{Lin
  et~al\mbox{.}}{2017b}]%
        {focal-loss}
\bibfield{author}{\bibinfo{person}{Tsung-Yi Lin}, \bibinfo{person}{Priya
  Goyal}, \bibinfo{person}{Ross Girshick}, \bibinfo{person}{Kaiming He}, {and}
  \bibinfo{person}{Piotr Doll{\'a}r}.} \bibinfo{year}{2017}\natexlab{b}.
\newblock \showarticletitle{Focal loss for dense object detection}. In
  \bibinfo{booktitle}{\emph{Proceedings of the IEEE international conference on
  computer vision}}. \bibinfo{pages}{2980--2988}.
\newblock


\bibitem[\protect\citeauthoryear{Lin, Maire, Belongie, Hays, Perona, Ramanan,
  Doll{\'a}r, and Zitnick}{Lin et~al\mbox{.}}{2014}]%
        {coco}
\bibfield{author}{\bibinfo{person}{Tsung-Yi Lin}, \bibinfo{person}{Michael
  Maire}, \bibinfo{person}{Serge Belongie}, \bibinfo{person}{James Hays},
  \bibinfo{person}{Pietro Perona}, \bibinfo{person}{Deva Ramanan},
  \bibinfo{person}{Piotr Doll{\'a}r}, {and} \bibinfo{person}{C~Lawrence
  Zitnick}.} \bibinfo{year}{2014}\natexlab{}.
\newblock \showarticletitle{Microsoft coco: Common objects in context}. In
  \bibinfo{booktitle}{\emph{European conference on computer vision}}. Springer,
  \bibinfo{pages}{740--755}.
\newblock


\bibitem[\protect\citeauthoryear{Liu, Huang, and Wang}{Liu
  et~al\mbox{.}}{2019}]%
        {adaptive-nms}
\bibfield{author}{\bibinfo{person}{Songtao Liu}, \bibinfo{person}{Di Huang},
  {and} \bibinfo{person}{Yunhong Wang}.} \bibinfo{year}{2019}\natexlab{}.
\newblock \showarticletitle{Adaptive nms: Refining pedestrian detection in a
  crowd}. In \bibinfo{booktitle}{\emph{Proceedings of the IEEE Conference on
  Computer Vision and Pattern Recognition}}. \bibinfo{pages}{6459--6468}.
\newblock


\bibitem[\protect\citeauthoryear{Liu, Qi, Qin, Shi, and Jia}{Liu
  et~al\mbox{.}}{2018b}]%
        {panet}
\bibfield{author}{\bibinfo{person}{Shu Liu}, \bibinfo{person}{Lu Qi},
  \bibinfo{person}{Haifang Qin}, \bibinfo{person}{Jianping Shi}, {and}
  \bibinfo{person}{Jiaya Jia}.} \bibinfo{year}{2018}\natexlab{b}.
\newblock \showarticletitle{Path Aggregation Network for Instance
  Segmentation}. In \bibinfo{booktitle}{\emph{Proceedings of IEEE Conference on
  Computer Vision and Pattern Recognition (CVPR)}}.
\newblock


\bibitem[\protect\citeauthoryear{Liu, Anguelov, Erhan, Szegedy, Reed, Fu, and
  Berg}{Liu et~al\mbox{.}}{2016}]%
        {ssd}
\bibfield{author}{\bibinfo{person}{Wei Liu}, \bibinfo{person}{Dragomir
  Anguelov}, \bibinfo{person}{Dumitru Erhan}, \bibinfo{person}{Christian
  Szegedy}, \bibinfo{person}{Scott Reed}, \bibinfo{person}{Cheng-Yang Fu},
  {and} \bibinfo{person}{Alexander~C Berg}.} \bibinfo{year}{2016}\natexlab{}.
\newblock \showarticletitle{Ssd: Single shot multibox detector}. In
  \bibinfo{booktitle}{\emph{European conference on computer vision}}. Springer,
  \bibinfo{pages}{21--37}.
\newblock


\bibitem[\protect\citeauthoryear{Liu, Liao, Hu, Liang, and Chen}{Liu
  et~al\mbox{.}}{2018a}]%
        {alf}
\bibfield{author}{\bibinfo{person}{Wei Liu}, \bibinfo{person}{Shengcai Liao},
  \bibinfo{person}{Weidong Hu}, \bibinfo{person}{Xuezhi Liang}, {and}
  \bibinfo{person}{Xiao Chen}.} \bibinfo{year}{2018}\natexlab{a}.
\newblock \showarticletitle{Learning efficient single-stage pedestrian
  detectors by asymptotic localization fitting}. In
  \bibinfo{booktitle}{\emph{Proceedings of the European Conference on Computer
  Vision (ECCV)}}. \bibinfo{pages}{618--634}.
\newblock


\bibitem[\protect\citeauthoryear{Pang, Xie, Khan, Anwer, Khan, and Shao}{Pang
  et~al\mbox{.}}{2019}]%
        {mgan}
\bibfield{author}{\bibinfo{person}{Yanwei Pang}, \bibinfo{person}{Jin Xie},
  \bibinfo{person}{Muhammad~Haris Khan}, \bibinfo{person}{Rao~Muhammad Anwer},
  \bibinfo{person}{Fahad~Shahbaz Khan}, {and} \bibinfo{person}{Ling Shao}.}
  \bibinfo{year}{2019}\natexlab{}.
\newblock \showarticletitle{Mask-guided attention network for occluded
  pedestrian detection}. In \bibinfo{booktitle}{\emph{Proceedings of the IEEE
  International Conference on Computer Vision}}. \bibinfo{pages}{4967--4975}.
\newblock


\bibitem[\protect\citeauthoryear{Redmon, Divvala, Girshick, and Farhadi}{Redmon
  et~al\mbox{.}}{2016}]%
        {yolo-v1}
\bibfield{author}{\bibinfo{person}{Joseph Redmon}, \bibinfo{person}{Santosh
  Divvala}, \bibinfo{person}{Ross Girshick}, {and} \bibinfo{person}{Ali
  Farhadi}.} \bibinfo{year}{2016}\natexlab{}.
\newblock \showarticletitle{You only look once: Unified, real-time object
  detection}. In \bibinfo{booktitle}{\emph{Proceedings of the IEEE conference
  on computer vision and pattern recognition}}. \bibinfo{pages}{779--788}.
\newblock


\bibitem[\protect\citeauthoryear{Redmon and Farhadi}{Redmon and
  Farhadi}{2017}]%
        {yolo-v2}
\bibfield{author}{\bibinfo{person}{Joseph Redmon} {and} \bibinfo{person}{Ali
  Farhadi}.} \bibinfo{year}{2017}\natexlab{}.
\newblock \showarticletitle{YOLO9000: better, faster, stronger}. In
  \bibinfo{booktitle}{\emph{Proceedings of the IEEE conference on computer
  vision and pattern recognition}}. \bibinfo{pages}{7263--7271}.
\newblock


\bibitem[\protect\citeauthoryear{Redmon and Farhadi}{Redmon and
  Farhadi}{2018}]%
        {yolo-v3}
\bibfield{author}{\bibinfo{person}{Joseph Redmon} {and} \bibinfo{person}{Ali
  Farhadi}.} \bibinfo{year}{2018}\natexlab{}.
\newblock \showarticletitle{Yolov3: An incremental improvement}.
\newblock \bibinfo{journal}{\emph{arXiv preprint arXiv:1804.02767}}
  (\bibinfo{year}{2018}).
\newblock


\bibitem[\protect\citeauthoryear{Ren, He, Girshick, and Sun}{Ren
  et~al\mbox{.}}{2015}]%
        {faster-rcnn}
\bibfield{author}{\bibinfo{person}{Shaoqing Ren}, \bibinfo{person}{Kaiming He},
  \bibinfo{person}{Ross Girshick}, {and} \bibinfo{person}{Jian Sun}.}
  \bibinfo{year}{2015}\natexlab{}.
\newblock \showarticletitle{Faster r-cnn: Towards real-time object detection
  with region proposal networks}. In \bibinfo{booktitle}{\emph{Advances in
  neural information processing systems}}. \bibinfo{pages}{91--99}.
\newblock


\bibitem[\protect\citeauthoryear{Russakovsky, Deng, Su, Krause, Satheesh, Ma,
  Huang, Karpathy, Khosla, Bernstein, et~al\mbox{.}}{Russakovsky
  et~al\mbox{.}}{2015}]%
        {imagenet}
\bibfield{author}{\bibinfo{person}{Olga Russakovsky}, \bibinfo{person}{Jia
  Deng}, \bibinfo{person}{Hao Su}, \bibinfo{person}{Jonathan Krause},
  \bibinfo{person}{Sanjeev Satheesh}, \bibinfo{person}{Sean Ma},
  \bibinfo{person}{Zhiheng Huang}, \bibinfo{person}{Andrej Karpathy},
  \bibinfo{person}{Aditya Khosla}, \bibinfo{person}{Michael Bernstein},
  {et~al\mbox{.}}} \bibinfo{year}{2015}\natexlab{}.
\newblock \showarticletitle{Imagenet large scale visual recognition challenge}.
\newblock \bibinfo{journal}{\emph{International journal of computer vision}}
  \bibinfo{volume}{115}, \bibinfo{number}{3} (\bibinfo{year}{2015}),
  \bibinfo{pages}{211--252}.
\newblock


\bibitem[\protect\citeauthoryear{Shao, Zhao, Li, Xiao, Yu, Zhang, and Sun}{Shao
  et~al\mbox{.}}{2018}]%
        {crowdhuman}
\bibfield{author}{\bibinfo{person}{Shuai Shao}, \bibinfo{person}{Zijian Zhao},
  \bibinfo{person}{Boxun Li}, \bibinfo{person}{Tete Xiao},
  \bibinfo{person}{Gang Yu}, \bibinfo{person}{Xiangyu Zhang}, {and}
  \bibinfo{person}{Jian Sun}.} \bibinfo{year}{2018}\natexlab{}.
\newblock \showarticletitle{Crowdhuman: A benchmark for detecting human in a
  crowd}.
\newblock \bibinfo{journal}{\emph{arXiv preprint arXiv:1805.00123}}
  (\bibinfo{year}{2018}).
\newblock


\bibitem[\protect\citeauthoryear{Song, Sun, Xie, Sun, and Pu}{Song
  et~al\mbox{.}}{2018}]%
        {tll}
\bibfield{author}{\bibinfo{person}{Tao Song}, \bibinfo{person}{Leiyu Sun},
  \bibinfo{person}{Di Xie}, \bibinfo{person}{Haiming Sun}, {and}
  \bibinfo{person}{Shiliang Pu}.} \bibinfo{year}{2018}\natexlab{}.
\newblock \showarticletitle{Small-scale pedestrian detection based on
  topological line localization and temporal feature aggregation}. In
  \bibinfo{booktitle}{\emph{Proceedings of the European Conference on Computer
  Vision (ECCV)}}. \bibinfo{pages}{536--551}.
\newblock


\bibitem[\protect\citeauthoryear{Wang, Chen, Yang, Loy, and Lin}{Wang
  et~al\mbox{.}}{2019}]%
        {guidedanchor}
\bibfield{author}{\bibinfo{person}{Jiaqi Wang}, \bibinfo{person}{Kai Chen},
  \bibinfo{person}{Shuo Yang}, \bibinfo{person}{Chen~Change Loy}, {and}
  \bibinfo{person}{Dahua Lin}.} \bibinfo{year}{2019}\natexlab{}.
\newblock \showarticletitle{Region proposal by guided anchoring}. In
  \bibinfo{booktitle}{\emph{Proceedings of the IEEE Conference on Computer
  Vision and Pattern Recognition}}. \bibinfo{pages}{2965--2974}.
\newblock


\bibitem[\protect\citeauthoryear{Wang, Xiao, Jiang, Shao, Sun, and Shen}{Wang
  et~al\mbox{.}}{2018}]%
        {repulsion-loss}
\bibfield{author}{\bibinfo{person}{Xinlong Wang}, \bibinfo{person}{Tete Xiao},
  \bibinfo{person}{Yuning Jiang}, \bibinfo{person}{Shuai Shao},
  \bibinfo{person}{Jian Sun}, {and} \bibinfo{person}{Chunhua Shen}.}
  \bibinfo{year}{2018}\natexlab{}.
\newblock \showarticletitle{Repulsion loss: Detecting pedestrians in a crowd}.
  In \bibinfo{booktitle}{\emph{Proceedings of the IEEE Conference on Computer
  Vision and Pattern Recognition}}. \bibinfo{pages}{7774--7783}.
\newblock


\bibitem[\protect\citeauthoryear{Yang, Zhang, Li, Zhang, and Sun}{Yang
  et~al\mbox{.}}{2018}]%
        {metaanchor}
\bibfield{author}{\bibinfo{person}{Tong Yang}, \bibinfo{person}{Xiangyu Zhang},
  \bibinfo{person}{Zeming Li}, \bibinfo{person}{Wenqiang Zhang}, {and}
  \bibinfo{person}{Jian Sun}.} \bibinfo{year}{2018}\natexlab{}.
\newblock \showarticletitle{Metaanchor: Learning to detect objects with
  customized anchors}. In \bibinfo{booktitle}{\emph{Advances in Neural
  Information Processing Systems}}. \bibinfo{pages}{320--330}.
\newblock


\bibitem[\protect\citeauthoryear{Zhang, Xiong, Sun, Hu, Li, and Yu}{Zhang
  et~al\mbox{.}}{2019}]%
        {doubleanchor}
\bibfield{author}{\bibinfo{person}{Kevin Zhang}, \bibinfo{person}{Feng Xiong},
  \bibinfo{person}{Peize Sun}, \bibinfo{person}{Li Hu}, \bibinfo{person}{Boxun
  Li}, {and} \bibinfo{person}{Gang Yu}.} \bibinfo{year}{2019}\natexlab{}.
\newblock \showarticletitle{Double Anchor R-CNN for Human Detection in a
  Crowd}.
\newblock \bibinfo{journal}{\emph{arXiv preprint arXiv:1909.09998}}
  (\bibinfo{year}{2019}).
\newblock


\bibitem[\protect\citeauthoryear{Zhang, Benenson, and Schiele}{Zhang
  et~al\mbox{.}}{2017}]%
        {citypersons}
\bibfield{author}{\bibinfo{person}{Shanshan Zhang}, \bibinfo{person}{Rodrigo
  Benenson}, {and} \bibinfo{person}{Bernt Schiele}.}
  \bibinfo{year}{2017}\natexlab{}.
\newblock \showarticletitle{Citypersons: A diverse dataset for pedestrian
  detection}. In \bibinfo{booktitle}{\emph{Proceedings of the IEEE Conference
  on Computer Vision and Pattern Recognition}}. \bibinfo{pages}{3213--3221}.
\newblock


\bibitem[\protect\citeauthoryear{Zhang, Wen, Bian, Lei, and Li}{Zhang
  et~al\mbox{.}}{2018}]%
        {orcnn}
\bibfield{author}{\bibinfo{person}{Shifeng Zhang}, \bibinfo{person}{Longyin
  Wen}, \bibinfo{person}{Xiao Bian}, \bibinfo{person}{Zhen Lei}, {and}
  \bibinfo{person}{Stan~Z Li}.} \bibinfo{year}{2018}\natexlab{}.
\newblock \showarticletitle{Occlusion-aware R-CNN: detecting pedestrians in a
  crowd}. In \bibinfo{booktitle}{\emph{Proceedings of the European Conference
  on Computer Vision (ECCV)}}. \bibinfo{pages}{637--653}.
\newblock


\bibitem[\protect\citeauthoryear{Zhou and Yuan}{Zhou and Yuan}{2018}]%
        {bibox}
\bibfield{author}{\bibinfo{person}{Chunluan Zhou} {and}
  \bibinfo{person}{Junsong Yuan}.} \bibinfo{year}{2018}\natexlab{}.
\newblock \showarticletitle{Bi-box regression for pedestrian detection and
  occlusion estimation}. In \bibinfo{booktitle}{\emph{ECCV}}.
  \bibinfo{pages}{135--151}.
\newblock


\bibitem[\protect\citeauthoryear{Zhu, Hu, Lin, and Dai}{Zhu
  et~al\mbox{.}}{2019}]%
        {deformablev2}
\bibfield{author}{\bibinfo{person}{Xizhou Zhu}, \bibinfo{person}{Han Hu},
  \bibinfo{person}{Stephen Lin}, {and} \bibinfo{person}{Jifeng Dai}.}
  \bibinfo{year}{2019}\natexlab{}.
\newblock \showarticletitle{Deformable convnets v2: More deformable, better
  results}. In \bibinfo{booktitle}{\emph{Proceedings of the IEEE Conference on
  Computer Vision and Pattern Recognition}}. \bibinfo{pages}{9308--9316}.
\newblock


\end{thebibliography}

\end{document}